\documentclass[hidelinks,onefignum,onetabnum]{siamart220329}
\newif\ifarxiv
\newif\ifsubmit

\arxivtrue   %<<< Uncomment as required
\submitfalse  %<<< Uncomment as required

\usepackage{lipsum}
\usepackage{amsfonts}
\usepackage{graphicx}
\usepackage{epstopdf}
\usepackage{algorithmic}
\ifpdf
  \DeclareGraphicsExtensions{.eps,.pdf,.png,.jpg}
\else
  \DeclareGraphicsExtensions{.eps}
\fi

\newsiamremark{remark}{Remark}
\headers{}{G. D. Er, S. Trimpe and M. Muehlebach}

\ifarxiv
\title{A Systems-Theoretic View on the Convergence of Algorithms under Disturbances}\fi
\ifsubmit
\title{A Systems-Theoretic View on the Convergence of Algorithms under Disturbances \thanks{Submitted to the editors, 19.12.2025.}}\fi
\ifarxiv
\author{Guner Dilsad Er \thanks{Max Planck Institute for Intelligent Systems, Tübingen, Germany
  (\email{gder@mpg.tue.de}, \email{michaelm@tue.mpg.de})}
 \and Sebastian Trimpe \thanks{Institute for Data Science in Mechanical Engineering, RWTH Aachen University, Aachen, Germany} (\email{trimpe@dsme.rwth-aachen.de})
\and Michael Muehlebach \footnotemark[1]}\fi

\ifsubmit
\author{Guner Dilsad Er \thanks{Max Planck Institute for Intelligent Systems, Tübingen, Germany
  (\email{gder@mpg.tue.de}, \email{michaelm@tue.mpg.de})}
 \and Sebastian Trimpe \thanks{Institute for Data Science in Mechanical Engineering, RWTH Aachen University, Aachen, Germany} (\email{trimpe@dsme.rwth-aachen.de})
\and Michael Muehlebach \footnotemark[2]}\fi

\usepackage{amsopn}

\usepackage{enumitem}
\usepackage{wrapfig}

\newcommand{\dm}[0]{\mathrm{d}\!}

\newtheorem{assumption}[theorem]{Assumption}
\newtheorem{example}[theorem]{Example}

\usepackage{pgfplots}
\usepackage{tikz}
\pgfplotsset{compat=1.18} % Specify the compatibility mode
\usepackage{tikz}
\usetikzlibrary{plotmarks}
\usetikzlibrary{arrows.meta,bending,positioning,calc} 
\usetikzlibrary[shapes,arrows]
\usepgfplotslibrary{fillbetween}
\usetikzlibrary{shapes.geometric, arrows}

\tikzstyle{startstop} = [rectangle, rounded corners, minimum width=3cm, minimum height=1cm,text centered, draw=black, fill=red!30]
\tikzstyle{process} = [rectangle, minimum width=3cm, minimum height=1cm, text centered, draw=black, fill=blue!30]
\tikzstyle{decision} = [diamond, minimum width=3cm, minimum height=1cm, text centered, draw=black, fill=green!30]
\tikzstyle{arrow} = [thick,->, >=stealth]

\begin{document}

\maketitle

% REQUIRED
\begin{abstract}
Algorithms increasingly operate within complex physical, social, and engineering systems where they are exposed to disturbances, noise, and interconnections with other dynamical systems.
This article extends known convergence guarantees of an algorithm operating in isolation (i.e., without disturbances) and systematically derives stability bounds and convergence rates in the presence of such disturbances. By leveraging converse Lyapunov theorems, we derive key inequalities that quantify the impact of disturbances. We further demonstrate how our result can be utilized to assess the effects of disturbances on algorithmic performance in a wide variety of applications, including communication constraints in distributed learning, sensitivity in machine learning generalization, and intentional noise injection for privacy. This underpins the role of our result as a unifying tool for algorithm analysis in the presence of noise, disturbances, and interconnections with other dynamical systems.
\end{abstract}

% REQUIRED
\ifsubmit
\begin{keywords}
Iterative algorithms, systems theory%, Lyapunov stability, converse Lyapunov theory, convergence analysis, optimization under disturbances, distributed learning, privacy-preserving learning, algorithmic stability
\end{keywords}

% REQUIRED
\begin{MSCcodes}
93D05, 93D30, 37N40 
%93D05 Lyapunov and other classical stabilities (Lagrange, Poisson, $L^p, l^p$, etc.) in control theory
%93D30 Lyapunov and storage functions
%37N40 Dynamical systems in optimization and economics 
\end{MSCcodes}\fi

\section{Introduction}

Algorithms increasingly operate in dynamic, uncertain environments where they are interconnected with physical, social, economic, or engineering systems. There is therefore a need for understanding the behavior of algorithms under disturbances, which may originate from various sources: communication delays in distributed systems, sensor noise on input data, quantization and finite-precision effects in digital implementations, approximation errors arising from simplified mathematical models, and even deliberate design choices such as noise injection for privacy and reduced precision for compute efficiency. These disturbances often reflect the reality of algorithm operation, where algorithms interact continuously with data streams and surrounding processes under real-time constraints. 

This article builds a rigorous foundation for analyzing iterative algorithms under such disturbances and provides corresponding convergence guarantees.
Our contributions are twofold: we derive convergence results for algorithms under bounded disturbances, and introduce new inequalities that explicitly capture the trade-offs between disturbance magnitude and convergence. This shifts the perspective from idealized, isolated algorithms to a more realistic view where algorithms are open systems embedded in noisy, resource-constrained environments.

The practical implications of our main result extend to several important application domains, which we briefly preview below and explore in depth in Section~\ref{sec:applications}. These examples represent three fundamentally different disturbance modeling paradigms: disturbances arising from algorithmic design choices, disturbances representing sensitivity with respect to input data, and deliberately introduced disturbances for functional purposes, such as promoting privacy.

\textbf{Application: Distributed Learning.} Distributed learning enables machine learning model training across multiple nodes without centralizing data, offering scalability and privacy. However, this design approach involves communication overhead, prompting the use of techniques that reduce communication frequency and volume \cite{Cao_2023, Er_2024,  FedPAQ_2020}. These techniques introduce delays, asynchrony, intermittent updates, packet loss, quantization errors, compression, message dropping, variable communication intervals, and node failures, all of which can be modeled as disturbances to the underlying optimization process. Our results provide a principled approach to analyzing how these disturbances impact the convergence of distributed optimization algorithms. In particular, our bounds offer insights into the trade-off between communication efficiency and convergence guarantees, guiding the design of robust, communication-efficient optimization methods. 

\textbf{Application: Algorithmic Stability and Generalization.} Stability, in the sense of sensitivity to perturbations of input data, is tightly linked to generalization in machine learning, as  algorithms that are less sensitive to changes in training data typically generalize better to unseen data \cite{Hardt_2016}. Our analysis revisits this connection: we model the influence of a single data point as a bounded disturbance, study how bounded disturbances affect algorithmic stability, and derive explicit bounds with our methodology. While previous work has highlighted this link under contraction-based assumptions \cite{Kozachkov_2023}, we take it a step further by relying solely on convergence and continuity properties that are inherent to gradient-based optimization algorithms. This provides generalization bounds that hold under weaker conditions than contraction. 

\textbf{Application: Privacy-Preserving Learning.} In privacy-preserving optimization, disturbances are deliberately introduced to protect sensitive data \cite{andrew_2021_dp, chaudhuri_2011_dp}. The challenge is to understand how much noise can be tolerated before the learning process breaks down. By treating privacy mechanisms as structured disturbances, our method provides a tool for navigating the trade-off between privacy and performance.

In addition to these applications, which we cover in detail in Section~\ref{sec:applications}, our analysis is also effective at capturing algorithms that influence the environments in which they operate, as discussed in Section~\ref{subsec:InterconnectedSystems}. These feedback effects are common in online learning, control, reinforcement learning, and performative prediction, where algorithm outputs shape the data or signals received by the environment. Our converse Lyapunov theorem provides convergence guarantees in these situations, despite the fact that disturbances are not bounded a priori.

\subsection{Related Work} Our work draws analogies between algorithms and dynamical systems, and builds on a recent trend in the literature \cite{Dörfler_2024,Er_2024,muehlebach_icml,Muehlebach_2020,su_differential_2016,Tong_2023,jordan_variational_2016}.  
Early works, such as \cite{Brockett_1991}, demonstrated how algorithms for sorting, diagonalizing matrices, and solving linear optimization problems can be modeled via continuous-time dynamical systems. This idea was further developed by framing a range of matrix analysis and decomposition problems within a differential geometric and dynamical systems setting \cite{Helmke_2012}. Analogies between algorithms and dynamical systems have contributed to the understanding of the long-term behavior of numerical integration methods \cite{Stuart_1998}, error propagation in iterative methods \cite{Kashima_2007}, and frequently inspire new algorithms; see, e.g., \cite{Attouch1,Attouch,Attouch2}. In optimization, insights from differential and symplectic geometry have greatly contributed to the analysis of momentum-based optimization algorithms \cite{su_differential_2016,jordan_variational_2016,Muehlebach_2020}, and led to new algorithms for constrained optimization \cite{JMLR:v23:21-0798,Mathprog}, distributed optimization \cite{Cummins_2025,Er_2024}, and variational inequalities \cite{zhang2025}. Another avenue of research has been to leverage integral quadratic constraints, a tool rooted in control theory, for the analysis of first-order optimization algorithms \cite{Lessard_2016,scherer2021convex}, which led to non-asymptotic convergence results for the alternating direction method of multipliers \cite{Attouch4,Nishihara_2015,Attouch3}, the introduction of the triple momentum method \cite{van2017fastest} (a very fast gradient-based optimizer on strongly convex functions), and the analysis of online optimization algorithms \cite{jakob2025}. This analysis and design paradigm has also been very important for the analysis of sampling algorithms \cite{vempala2019rapid,wibisono2018sampling,guo2024provable}, generative modeling \cite{song2020score,berner2022optimal}, minimax algorithms, algorithms evolving on manifolds \cite{schechtman2023orthogonal, absil2009optimization}, and many more. 

While Lyapunov-based methods are well-established in the analysis of dynamical systems, their use in algorithmic contexts has been limited to constructing explicit Lyapunov functions to certify convergence. In contrast, our work introduces \emph{converse Lyapunov theorems} \cite{hahn1967stability,Sastry_1999,Khalil_2002nonlinear} for the analysis of algorithms -- a perspective that, to our knowledge, is fundamentally novel within the computer science and algorithmic literature. From this standpoint, the algorithm is treated as a dynamical system whose stability properties imply the existence of a Lyapunov function, which can then be systematically exploited to analyze the effects of disturbances. This shift in perspective provides a principled and general result for deriving robustness guarantees and convergence bounds for perturbed algorithms, and as a consequence, our approach opens new avenues for studying algorithm performance in noisy, interconnected, or uncertain environments, as highlighted with numerous examples. In addition, our work significantly extends converse Lyapunov results from dynamical systems theory in order to provide results that are practically relevant for algorithm design. More precisely, we construct Lyapunov functions that are able to preserve convergence rates and resort to a pseudometric instead of a metric to quantify convergence. The latter is important in the context of non-convex optimization, for example, where algorithms typically converge to the set of critical points (disconnected) instead of a single equilibrium.

There are numerous related works concerning each of the examples that will be discussed in the following. These works will be highlighted in the corresponding subsections.

\textbf{Outline:} The article is organized as follows. In Section~\ref{sec:prelim}, we introduce the problem setup and define the class of algorithms considered in our analysis. We also formalize key assumptions and stability definitions using pseudometrics. Our main results are presented in Section~\ref{sec:main}, where we establish a convergence result for an algorithm operating under disturbances by leveraging converse Lyapunov arguments. These results characterize how convergence properties of the algorithm operating in isolation lead to guarantees under disturbances. Section~\ref{sec:proof} contains the detailed proofs of our main results. In Section~\ref{sec:applications}, we demonstrate how our results apply to three central areas in algorithm design: communication-efficient distributed optimization, algorithmic stability and generalization, and privacy-preserving learning. These examples show that our analysis recovers known results in the literature and provides insights into the design and evaluation of learning algorithms under noise and structural perturbations. Finally, Section~\ref{sec:conclusions} summarizes our findings.

\section{Preliminaries and Problem Setup}\label{sec:prelim}

\subsection{Definitions}

We model an algorithm as a discrete-time dynamical system with the state $x_k \in \mathbb{R}^d$. The evolution of the algorithm's state over the iterations is described by the discrete-time dynamics
\begin{align}
    x_{k+1} = f_k( x_k), \label{eqn:nominal_system_dynamics}
\end{align}
which specify the algorithm's evolution when operated in isolation. The central theme of the article is to analyze the algorithm's evolution when subjected to disturbances or interacting with surrounding dynamic processes. This is modeled via the disturbed discrete-time dynamics
\begin{align}
    z_{k+1} = g_k( z_k, e_k), \label{eqn:disturbed_system}
\end{align}
where $ z_k  \in \mathbb{R}^d $ is the algorithm's state in the presence of disturbances, $ e_k $ are the disturbances, and $ g_k $ satisfies $ g_k( z_k, 0) = f_k(z_k)$. As we present in the next sections, the disturbances may arise from bounded deterministic mechanisms (Theorem~\ref{thm:mainthm}), zero-mean stochastic noise (Theorem~\ref{thm:zero_mean_stochastic}), or state-dependent processes; that is, $e_{k+1}=\Delta_k(e_k,z_k)$, where $\Delta_k$ denotes the dynamics of the disturbance-generating process (Corollary~\ref{corr:interconnected}). 

We impose the following basic assumptions for both nominal and disturbed system dynamics.

\begin{assumption}\label{ass:uni_Lipschitz_f} Let $|\cdot|: \mathbb{R}^d \to \mathbb{R}_{\geq 0}$ be any norm on $\mathbb{R}^d$. 
The nominal dynamics $f_k(x)$ are Lipschitz in $x$ uniformly with respect to $k$, that is, there exists a constant  $L_f$ satisfying
$$| f_k\left(x_1\right)-f_k\left( x_2\right)|\leq L_f| x_1-x_2 |,$$ for all $k\geq 0$ and any $x_1, x_2 \in \mathbb{R}^d$.
\end{assumption}

\begin{assumption} \label{ass:bounded_effect_g} 
    The disturbed dynamics $g_k(x,e)$ are {locally} Lipschitz in $e$ with a time-dependent Lipschitz constant $L_{\mathrm{e}k}$, satisfying
     $$| g_k\left(x, e_1\right)-g_k\left(x, e_2\right)|\leq L_{\mathrm{e}k}| e_1-e_2 |,$$ for any $e_1$, $e_2$ in a compact set $\Omega$ that contains the origin.
\end{assumption}
%=======%
These assumptions impose basic regularity properties on the dynamics. Both Assumption~\ref{ass:uni_Lipschitz_f} and~\ref{ass:bounded_effect_g} hold for a wide range of iterative algorithms, including optimization algorithms based on gradient descent and variants thereof.

However, these assumptions may fail to hold for algorithms that include combinatorial selection rules such as top-$k$ sparsification \cite{Stich_2018} or quantized updates \cite{Alistarh_2017}.
These elements can introduce discontinuities or unbounded sensitivities in $g_k$ with respect to $e$ and violate our assumptions. In particular, the top-$k$ operator exhibits discontinuous behavior when multiple entries have similar magnitudes, while hard thresholding results in abrupt changes in the update dynamics based on fixed cutoffs. Similarly, quantization schemes involve piecewise constant mappings, which are not Lipschitz continuous and may amplify small perturbations in discontinuous ways. In these cases, the non-smooth operation can be approximated by a smooth counterpart, while the remaining non-smooth effect is captured through the disturbance term $e_k$. Under this interpretation, Assumptions~\ref{ass:uni_Lipschitz_f} and~\ref{ass:bounded_effect_g} continue to hold.
%=======%

To characterize the flow of the nominal algorithm, i.e., the evolution of iterates under varying initial conditions, we define the operator $ \phi(k', k,\xi)$:
\begin{align*}
     \phi(k', k, \xi)= f_{k+k'-1}\circ\ldots \circ f_{k+1} \circ f_k(\xi),
\end{align*} 
which describes the iterate after $k'$ iterations, starting from the initial condition $\xi$ at time $k$.
The flow map $\phi$ allows us to analyze how the initial state $ \xi $ evolves through the nominal algorithm dynamics over $k'$ iterations. By the composition of Lipschitz continuous functions, the flow map $ \phi(k',k, \xi) $ becomes Lipschitz continuous in $\xi$, i.e.,
\begin{align}\label{eqn:L_cont_phi_e}
\left| \phi(k', k, \xi_1) - \phi(k', k,  \xi_2) \right| \leq L^{k'}_f \left| \xi_1 - \xi_2 \right|, \quad \forall k'\geq 0, \quad \forall k\geq 0.
\end{align}

\subsection{Nominal Convergence of Algorithms}

Stability and convergence are interrelated concepts crucial for understanding the behavior of algorithms. Stability characterizes the sensitivity of the system to initial conditions and perturbations, while convergence describes the tendency of the iterates of an algorithm to approach an optimal solution over time. 

We start by defining a pseudometric that serves as a generalized distance measure for capturing different notions of convergence, such as closeness in parameter space ($|x-x_\ast|$) or decrease in objective function values ($\ell(x)-\ell(x_\ast)$).

\begin{definition}\label{defn:metric_defn} 
The function $ \dm: \mathbb{R}^d\times \mathbb{R}^d \rightarrow \mathbb{R}_{\geq 0}$ is a pseudometric on $\mathbb{R}^d$ if it satisfies,
\begin{align*}
\dm\left(x,x\right) = 0, \quad
\dm\left(x,y\right)=\dm\left(y,x\right),\quad \text{and}\quad
\dm\left(x,z\right)\leq\dm\left(x,y\right)+\dm\left(y,z\right),
\end{align*}
for all $x,y,z\in \mathbb{R}^d$. Additionally, $\dm$ is assumed to be norm-bounded,
\begin{align}
  \dm\left(x,y\right) \leq L_\dm~|x-y|.  \label{eqn:metric_norm_bound}
\end{align}\end{definition}

The following assumption characterizes the convergence and stability of the nominal algorithm dynamics.
\begin{assumption}\label{ass:convergence}
There exist a constant $c_0 > 0$ and a nondecreasing sequence ${\tau(i)}$ for all $i\geq0$ with $\tau(i)\leq \tau(i+1)$ such that, for all $k\geq 0$ and $k' \geq 0$,
\begin{align}
\dm\left(\phi(k',k, \xi),x_\ast\right) \leq c_0 \left( \prod_{i=k}^{k+k'-1} \tau(i)\right) \dm\left(\xi,x_\ast\right),\quad \forall \xi \in \mathbb{R}^d, \label{eqn:stability}\end{align}
where $\dm$ denotes any pseudometric defined in Definition~\ref{defn:metric_defn} and the multiplicative factor $\left( \prod_{i=k}^{k+k'\!-1} \!\tau(i)\!\right) $ characterizes how quickly the system converges to an equilibrium $x_\ast$.
Moreover, there exists an integer $K>0$ such that, for all $k\geq 0$ and all $k' \geq K$,
\begin{align}
\dm\left(\phi(k',k, \xi),x_\ast\right) \leq \left( \prod_{i=k}^{k+k'-1} \tau(i)\right) \dm\left(\xi,x_\ast\right),\quad \forall \xi \in \mathbb{R}^d.\label{eqn:ass_2}\end{align}\end{assumption}

The formulation via \eqref{eqn:stability} accommodates a broad set of commonly used convergence measures, including function values and gradients in optimization algorithms. We do not restrict ourselves to linear convergence, and, in fact, \eqref{eqn:stability} can accommodate all the typical sublinear convergence rates, such as $1/k$, $1/\sqrt{k}$, etc., as long as the strengthened condition \eqref{eqn:ass_2} is satisfied.

The following example provides a canonical algorithm that is modeled by \eqref{eqn:disturbed_system} and satisfies Assumptions~\ref{ass:uni_Lipschitz_f} and \ref{ass:convergence}. 

\begin{example}\label{ex:SGD} % Concrete Example
Let $\ell$ be an $L_\ell-$Lipschitz, $\beta-$smooth, and convex loss function that is minimized through gradient descent with additive noise,
\begin{align*}
    x_{k+1}=x_k-h\left(\nabla\ell(x_k)+n_k\right), \quad n_k \overset{i.i.d.}{\sim} \mathcal{N}(0, \sigma_n^2 \mathrm{I}_d/d),
\end{align*}
where $h={1}/{\beta}$ is the step size and $n_k$ is the noise sampled from a normal distribution with zero mean and variance of $\sigma_n^2$. In the absence of noise, we know from \cite{Nesterov_2018} that gradient descent generates a sequence $x_k$, $k=0,1,\dots,$ satisfying the following inequality, 
\begin{align*}
    \ell(x_k)-\ell(x_\ast)\leq \frac{2\beta}{k+4}\left|x_0-x_\ast\right|^2.
\end{align*} 
This relation is captured with \eqref{eqn:stability} by defining the pseudometric and associated rate function as 
\begin{align*}
\dm\left(x_k,x_\ast\right)=\ell(x_k)-\ell(x_\ast), \quad \prod_{i=0}^{k-1} \tau(i) =\frac{4}{k+4}, \quad \tau(i)=1-\frac{1}{i+5}.
\end{align*} 
The choice of $\dm$ satisfies the properties listed in Definition~\ref{defn:metric_defn} under the assumption that $\ell$ is $L_\ell-$Lipschitz. The rate $\tau$ captures the convergence of the nominal gradient-descent algorithm (with $n_k=0$). The noise enters linearly with gain $L_\mathrm{e} = h$, and its impact on convergence follows from Theorem~\ref{thm:zero_mean_stochastic}, which extends Theorem~\ref{thm:mainthm} to zero-mean stochastic disturbances.\end{example}

In the next section, our main result, Theorem~\ref{thm:mainthm}, provides a unifying framework for analyzing the convergence of algorithms under disturbances. Through the use of a carefully constructed Lyapunov function, Theorem~\ref{thm:mainthm} quantifies the effects of bounded disturbances on algorithmic performance, offers stability guarantees without requiring contractivity, and accommodates various convergence rates through the rate functions $\prod_{i=0}^{k-1} \tau(i)$. This approach effectively analyzes both contractive and non-contractive systems while providing explicit bounds on the impact of disturbances.

\section{Main Results}\label{sec:main}

In this section, we state and discuss the main result of this work, Theorem~\ref{thm:mainthm}, which quantifies how disturbances affect convergence. Its proof is deferred to Section~\ref{sec:proof}.

\begin{theorem}\label{thm:mainthm}
Let the unperturbed algorithm be represented by the nominal dynamics in \eqref{eqn:nominal_system_dynamics} and satisfy Assumptions~\ref{ass:uni_Lipschitz_f}, \ref{ass:bounded_effect_g}, and \ref{ass:convergence}. Then, there exists a constant $L_V>0$, such that the following bound holds for the perturbed algorithm dynamics, 
\begin{align}  \label{eqn:rate_matching_z}
    \dm\left(z_k,x_\ast\right) \leq c_0 \left(\prod_{i=0}^{k-1} \tau(i)\right)\dm\left(z_0,x_\ast\right) +L_V   \sum_{j=0}^{k-1}\left(\prod_{i=j+1}^{k-1} \tau(i)\right)L_{\mathrm{e}j}|e_{j}| ,
\end{align}
for all $z_0 \in \mathcal{S}$, where $S\subset \mathbb{R}^d$ is compact, $z_k$ denotes the state of the perturbed algorithm \eqref{eqn:disturbed_system}, $x_\ast$ is an equilibrium of the nominal dynamics \eqref{eqn:nominal_system_dynamics}, $|e_j|$ represents the disturbance bound, and the constant $L_{\mathrm{e}j}>0$ is defined in Assumption~\ref{ass:bounded_effect_g}. 
\end{theorem}

Theorem~\ref{thm:mainthm} establishes an important relationship between unperturbed and perturbed algorithm dynamics. 
Notably, the methodology developed here shows that stability in the sense of Assumption~\ref{ass:convergence} and continuity assumptions in Assumptions~\ref{ass:uni_Lipschitz_f} and~\ref{ass:bounded_effect_g} are sufficient to establish bounds on the effect of disturbances. 
This is achieved by constructing a Lyapunov function and showing that this Lyapunov function is uniformly Lipschitz continuous with respect to the algorithm's state.

The novelty of the result lies in the following aspects: i) The analysis avoids the much stronger notion of contraction introduced in earlier works \cite{Kozachkov_2023, Wensing_Slotine_2020}.
ii) The analysis provides a precise quantification of the convergence rate, while the literature on input-to-state stability and converse Lyapunov theory focuses on qualitative robustness guarantees such as asymptotic gain bounds \cite[Theorem 5.17]{Sastry_1999}.

The bound provided in Theorem~\ref{thm:mainthm} accommodates different classes of disturbances by applying Hölder's inequality, as illustrated in the following remark.
\begin{remark} \label{remark:holder} Applying Hölder's inequality to the bound in Theorem~\ref{thm:mainthm} separates the contributions of the disturbance signal $e_{j}$ and the rate term $\tau(i)$. This yields, for any $p,q\geq 1$ satisfying $\frac{1}{p}+\frac{1}{q}=1$, 
\begin{align*}
 \dm\left(z_k,x_\ast\right) \leq c_0\! \left(\prod_{i=0}^{k-1} \tau(i)\right)\!\dm\left(z_0,x_\ast\right) +L_V  {\left( \sum_{j=0}^{k-1}\left(L_{\mathrm{e}j}|e_{j}|\right)^p\right)}^{\!\!\frac{1}{p}} \left( \sum_{j=0}^{k-1}\left(\prod_{i=j+1}^{k-1} \tau(i)\right)^{\!\!q}\right)^{\!\!\frac{1}{q}}\!\!.
\end{align*}

Choosing $p=\infty$, $q=1$ yields a bound based on the maximum disturbance, while $p = q = 2$ captures its energy. The bound is particularly useful when analyzing specific types of disturbances, such as bounded, impulsive, or stochastic signals. 
\end{remark}

In summary, Theorem~\ref{thm:mainthm} characterizes the convergence of an algorithm when subjected to disturbances, based on the convergence rates corresponding to the operation in isolation. The result is also of interest to the dynamical systems and control community, as it hinges on a converse Lyapunov theorem that requires comparably weak assumptions (Lipschitz continuity of the dynamics).

\subsection{Extension to Zero-Mean Stochastic Disturbances} \label{sec:zero-mean}

The bound in Theorem~\ref{thm:mainthm} applies to bounded disturbances. In many practical scenarios, however, perturbations are stochastic, for example, due to additive zero-mean noise with bounded second moment. Such stochastic disturbances can be incorporated into our methodology by taking expectations with respect to the noise distribution. The result below requires a smoothness condition on the system flow and the disturbed dynamics, in addition to Assumptions~\ref{ass:uni_Lipschitz_f} and~\ref{ass:convergence}.

\begin{assumption}\label{ass:smooth_flow} The nominal dynamics $f_k(x)$ are twice continuously differentiable with respect to $x$, and the norm of their second derivative is bounded by a constant $L_{H_f}>0$, uniformly in $k$. 
\end{assumption}
As a consequence of Assumption~\ref{ass:smooth_flow}, for all $k'>0$, the flow map $\phi(k',k,\xi)$ is twice continuously differentiable with respect to $\xi$, and the norm of its second derivative is bounded by a constant, for any $\xi\in\mathcal{S}$ and any $k>0$.

\begin{assumption} \label{ass:vanish_hessian_g}
The disturbed dynamics $g_k(x,e)$ can be stated as
\begin{align*}
    g_k(x,e)=f_k(x)+A_ke,
\end{align*} for any $e \in \mathbb{R}^e$, where $f_k$ represents the nominal dynamics and $|A_k|\leq L_{\mathrm{e}k}$. In particular, this implies $|\frac{\partial  g_k(x,e)}{\partial e}|\leq L_{\mathrm{e}k}$ and $|\frac{\partial^2  g_k(x,e)}{\partial e^2}|=0$.
\end{assumption}

\begin{assumption}\label{ass:smooth_d} The distance function $\dm~$, defined according to Definition~\ref{defn:metric_defn}, is chosen such that $\dm~$ is twice continuously differentiable, and the norm of its Hessian is uniformly bounded by a constant $L_{H_\dm}>0$. 
\end{assumption}

Under these assumptions, the following theorem establishes the corresponding convergence guarantee.

\begin{theorem}\label{thm:zero_mean_stochastic}
Let the unperturbed algorithm be represented by the nominal dynamics in \eqref{eqn:nominal_system_dynamics}, and let Assumption~\ref{ass:uni_Lipschitz_f}, \ref{ass:convergence}, \ref{ass:smooth_flow}, \ref{ass:vanish_hessian_g}, and \ref{ass:smooth_d} hold. Then, there exists a constant $L_H>0$, such that the perturbed dynamics satisfy
\begin{align}\label{eqn:main_stochastic}
        \mathbb{E}\left\{\dm\left(z_k,x_\ast\right) \right\} \leq \bar{c}_0 \dm\left(z_0,x_\ast\right)\left( \prod_{i=0}^{k-1} \tau(i)\right)+\frac{1}{2}L_H\sigma_n^2\sum_{j=0}^{k-1}\left( \prod_{i=j+1}^{k-1}\tau(i) \right)L_{\mathrm{e}j}^2,
\end{align}
for all $z_0 \in \mathcal{S}$, where $S\subset \mathbb{R}^d$ is compact, $z_k$ denotes the state of the perturbed algorithm \eqref{eqn:disturbed_system}, $x_\ast$ is an equilibrium of the nominal dynamics \eqref{eqn:nominal_system_dynamics}, and $n_k$ represents zero-mean noise independent across time, with $\mathbb{E}\{|n_k|^2\}\leq\sigma_n^2$. The constants $L_f$ and $L_{\mathrm{e}k}$ are defined in Assumptions~\ref{ass:uni_Lipschitz_f} and~\ref{ass:vanish_hessian_g}, respectively. 
\end{theorem}

\begin{proof}
The result follows from Lemma~\ref{lemma:alternative_main}, which characterizes the behavior of the chosen Lyapunov function ${V}$ under disturbances. Unrolling the recursion in \eqref{eqn:V_tilde_pertdecrease_inlemma} gives
\begin{align}
        \mathbb{E}\left\{{V}(k,z_{k}) \right\} \leq{V}(0,z_0)\left( \prod_{i=0}^{k-1} \tau(i) \right)+\frac{1}{2}L_H\sigma_n^2\sum_{j=0}^{k-1}\left( \prod_{i=j+1}^{k-1} \tau(i) \right) L_{\mathrm{e}j}^2.\label{eqn:Vtilde_rec}
\end{align}
Substituting bounds in \eqref{eqn:Vtilde_bound} into \eqref{eqn:Vtilde_rec} yields the desired result \eqref{eqn:main_stochastic}.
\end{proof}

The bound in Theorem~\ref{thm:zero_mean_stochastic} holds for the last iterate (non-averaged) of the stochastic dynamics. Unlike many analyses that rely on iterate averaging to obtain expectation bounds, our methodology controls the last iterate under mild regularity assumptions similar to the deterministic case. This feature is both conceptually and practically important, as it provides interpretable guarantees without modifying the algorithm. We revisit this extension in Section~\ref{sec:dp}, where a concrete example demonstrates how our main theorem applies to a case with stochastic disturbances.

\subsection{Relation to Input-to-State Stability}

In this work, we analyze the convergence of algorithms under disturbances using a systems-theoretic approach. A related and well-established concept in systems theory is input-to-state stability (ISS), which provides a framework that quantifies how a system's state responds to external disturbances \cite{Sontag_book_2008}. ISS states that the system's state remains bounded and converges to a neighborhood of the equilibrium, where the bounds of the neighborhood depend on the cumulative effect of the disturbance.

To understand how ISS connects to our analysis, we now recall that ISS is equivalent to the combination of two properties: global asymptotic stability in the absence of disturbances (0-GAS) and a limit property that ensures the trajectories remain bounded and approach the invariant set with a margin determined by the disturbance magnitude.

This equivalence, although non-trivial, provides valuable insight into the connection between asymptotic stability and robustness under external inputs. We refer readers who are interested in the details of this equivalence to the seminal work \cite{Sontag_Wang_1996_new_char}.

In contrast to classical approaches \cite{Sontag_Wang_1996_new_char}, we derive ISS-like properties from the combination of stability properties of the unperturbed system (i.e., 0-GAS with explicit convergence rate), and Lipschitz continuity of the system dynamics with respect to disturbances.

%{LIM and Lipschitz continuity}
The limit property and Lipschitz continuity both concern how systems respond to disturbances, but they emphasize different aspects of that response. The limit property  ensures that trajectories remain ultimately bounded near an invariant set despite persistent inputs. This property is inherently global and directly tied to ISS. On the other hand, Lipschitz continuity refers to the local sensitivity of a system's response to disturbances. 
It ensures that the change in the system's state due to disturbances is bounded proportionally, which means that the system's trajectories do not change too abruptly in response to small variations in the disturbance input. 

Lipschitz continuity is relatively straightforward to verify since it only requires bounding the system's sensitivity to small input variations, often using Jacobian estimates. In contrast, establishing the limit property involves analyzing long-term trajectory behavior under disturbances, which typically requires Lyapunov methods or asymptotic arguments. This makes Lipschitz continuity easier to verify.

We also extend these insights to applications in machine learning and distributed optimization. Our approach serves as a blueprint for incorporating stability and robustness into the design and evaluation of optimization methods, particularly in applications involving distributed learning, privacy-preserving algorithms, or systems with inherent noise.

\subsection{Relation to Interconnected Systems and Small Gain Theorem}\label{subsec:InterconnectedSystems}
~ 
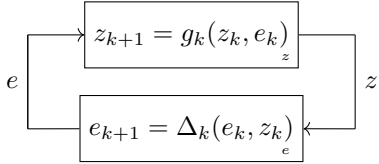
\begin{wrapfigure}{l}{0.5\textwidth}
    \centering
    \tikzstyle{block} = [draw, rectangle, minimum height=2.5em, minimum width=2.5em]
\tikzstyle{largeblock} = [draw, rectangle, minimum height=1em, minimum width=5em]
\tikzstyle{smallblock} = [rectangle, minimum height=0.5em, minimum width=0.5em]
\tikzstyle{longblock} = [draw, rectangle, minimum height=5em, minimum width=1em]

\tikzstyle{sum} = [draw, circle, node distance=2cm]
\tikzstyle{input} = [coordinate]
\tikzstyle{elbow} = [coordinate]

\tikzstyle{pinstyle} = [pin edge={to-,thin,black}]

\begin{tikzpicture}

% Blocks
\node [input](V) {};

\node [block, right= 0.75 cm of V] (linear) {$z_{k+1}=g_k(z_k,e_k)$};
\node at ([shift={(1.25cm,-0.3cm)}]$(linear)$) {\tiny{$z$}};

\node [elbow, right= 0.75 cm of linear](E1){};
\node [elbow, below= 1.25 cm of E1](E2){};
\node [elbow, below= 1.25 cm of V] (E4) {};

\node [block, below= 0.80 cm of $(linear)$] (nonlinear) {$e_{k+1}=\Delta_k(e_k,z_k)$};
\node at ([shift={(1.25cm,-0.3cm)}]$(nonlinear)$){\tiny{$e$}};

% Connections
\draw [-] (E4) -- node[midway, left] {$e$} (V);
\draw [->] (V) --  (linear);
\draw [-](linear) -- (E1);
\draw [-] (E1) -- node[midway, right] {$z$} (E2);
\draw [->](E2) -- (nonlinear);

\draw [-] (nonlinear) -- node[midway, below] {} (E4);
\draw [-] (E4) -- node[midway, below] {} (V);

\end{tikzpicture}
    \caption{Illustration of an interconnected system}
    \label{fig:interconnected_system}
\end{wrapfigure}

As algorithms do not operate in isolation, their updates often influence the environments in which they are operating. This naturally leads to a feedback structure, where the environment responds to algorithm behavior and, in turn, acts as a state-dependent disturbance. From a systems-theoretic perspective, this interaction can be modeled as an interconnected system, where the algorithm dynamics are coupled to a disturbance mechanism $\Delta_k(e,z)$, as depicted in Figure~\ref{fig:interconnected_system}.

In the control theory literature, the stability of such systems is studied using a notion of dissipation \cite{Sontag_Wang_1995}.
Suppose the disturbance dynamics in open-loop $e_{k+1}=\Delta_k(e_k,z_k)$, where $z_k$ is an exogenous disturbance, admit a function $V_\Delta: \mathbb{N} \times \mathbb{R}^{n_\text{e}}\rightarrow \mathbb{R}_{\geq 0}$ that satisfies the following dissipation inequality:
\begin{align}
    \label{eqn:V_Liss}     {V}_{\Delta}(k+1,\Delta_k(e,z))-{V}_{\Delta}(k,e)  \leq - a|e|+b|z|, \quad \forall k\geq 0,~\forall e \in \Omega,~\forall z \in \mathbb{R}^d,
\end{align}
for some positive constants $a,b>0$, where $a$ characterizes the decay rate of the initial state’s influence over time (dissipation), and $b$ is the disturbance gain (supply). 

The Lipschitz-based gain bounds we derive in Theorem~\ref{thm:mainthm} quantify how disturbances are amplified through the algorithm’s update dynamics. This amplification plays a role directly analogous to the gain functions in the small gain theorem \cite{Jiang_1994}, where stability of \emph{interconnected systems} is guaranteed when the product of subsystem gains remains below a threshold. We formalize this in the following corollary, which is an instance of the small-gain theorem.

\begin{corollary} 
\label{corr:interconnected} % closed loop stability based on Theorem~\ref{thm:main_thm}
Let the algorithm dynamics satisfy the assumptions of Theorem~\ref{thm:mainthm} and suppose the algorithm forms a feedback interconnection as shown in Figure~\ref{fig:interconnected_system} with the disturbance dynamics $e_{k+1}=\Delta_k(e_k,z_k)$. Let the disturbances $e_k$ satisfy the inequality \eqref{eqn:V_Liss}, where $V_\Delta$ is a positive definite function and $\dm~$ is a norm. Then, for sufficiently  small $L_V\sup_{k\in \mathbb{N}} L_{\mathrm{e}k}$, there exists a nonempty set of initial conditions $X$ with $0\in X$ and a constant $c_\mathrm{b}>0$ such that the trajectories of the interconnected system satisfy
\begin{equation*}
    |(z_k,e_k)| \leq c_\mathrm{b}, \quad \forall k\geq 0, \quad \lim_{k\rightarrow\infty} |(z_k,e_k)| = 0,
\end{equation*}
for all initial conditions $(z_0,e_0)\in X$.
\end{corollary} 
\begin{proof} We first fix an initial condition $(z_0,e_0)$ and define the sequence 
\begin{align*}
    W_{k+1}=V({k+1},z_{k+1})+m V_{\Delta}(k+1,e_{k+1}),
\end{align*}
for some constant $m>0$. From Lemma~\ref{lemma:V_to_Vpert}, we know that under continuity and convergence assumptions (Assumptions~\ref{ass:uni_Lipschitz_f} and \ref{ass:convergence}), the following inequality holds
\begin{align*}
    V(k+1,z_{k+1})- V(k,z_{k})\leq (\tau(k)-1)V(k,z_k)+ L_VL_{\mathrm{e}k}|e_k|. 
\end{align*} 
Moreover, the disturbance dynamics are assumed to satisfy \eqref{eqn:V_Liss}. Substituting $V$ and $V_\Delta$ into the difference of the composite Lyapunov function $W_{k+1}-W_{k}$, we obtain
\begin{align}
    W_{k+1}-W_{k}\leq (\tau(k)-1)V(k,z_k) + L_VL_{\mathrm{e}k}|e_k| +m \left(-a|e_k|+b|z_k|\right).
\end{align}
Hence, due to the fact that $L_V\sup_{k\in \mathbb{N}} L_{\mathrm{e}k}$ is sufficiently small, there exists a constant $m>0$ such that %for sufficiently small $L_V\sup_{k\in \mathbb{N}}L_{\mathrm{e}k}$, 
the right-hand side becomes strictly negative. This implies $W_k$ is strictly decreasing.
    As $W_{k}$ is nonnegative and strictly decreasing along trajectories of the interconnected system, it follows that the solutions $(z_k,e_k)$ are bounded for all $k>0$ and converge to the origin. This concludes the proof.
\end{proof}
%-----------------Performative Prediction-----------------%

Building this connection allows us to capture concepts such as \emph{performative prediction} \cite{Perdomo_2020} and \emph{decision dependence} \cite{Zhiyu_2025}. In performative prediction, predictions affect the environment in ways that transform future data distributions. This feedback can be captured through a response function $\Delta(z)$, which maps the algorithm state to the outcome it causes. Note that, in such cases, the aforementioned disturbance dynamics $\Delta_k(e,z)$ reduce to a static map of the current state of the algorithm, which is the standard assumption in performative prediction \cite{Perdomo_2020}. 

The analysis from \cite{Perdomo_2020} demonstrates that the stability of repeated risk minimization in performative settings depends on properties of the response function. 
A key insight from the work is that continuity of $\Delta$ is a sufficient condition for the existence of equilibria (fixed points) by Brouwer's fixed point theorem.
Similarly, under the Lipschitz continuity assumptions, our results in Theorem~\ref{thm:mainthm} and Corollary~\ref{corr:interconnected} suggest that such performative dynamics are stable, so that convergence to a fixed point is guaranteed when the feedback gain is sufficiently small.

In conclusion, performative prediction is a special case of our more general feedback interconnection model. Our analysis unifies such settings under a common perspective and extends them to dynamic disturbance processes.

\section{Proof of Theorem~\ref{thm:mainthm}}\label{sec:proof}
At the heart of Theorem~\ref{thm:mainthm} lies a converse Lyapunov result that constructs a Lyapunov function, assuming that the dynamics are stable in the sense of \eqref{eqn:stability} and satisfy Assumption~\ref{ass:uni_Lipschitz_f}. Compared to earlier results in the literature, such as \cite{Jiang_Wang_2002_conv}, \cite[ Theorem 4.14]{Khalil_2002nonlinear}, and \cite[ Theorem 5.17]{Sastry_1999}, our construction is novel in that it guarantees that the convergence rate is preserved. The converse Lyapunov theorem is summarized and proved below, and constitutes the first important step in the proof of Theorem~\ref{thm:mainthm}.

\begin{theorem} \label{thm:converselyapunov} Let Assumption~\ref{ass:convergence} hold.
Then, there exists a Lyapunov function $V:\mathbb{N}\times \mathbb{R}^d \rightarrow \mathbb{R}_{\geq 0}$,
\begin{align}
V(k,\xi) := \sup_{k' \geq 0}~\dm\left(\phi(k',k, \xi),x_\ast\right) \Phi(k, k'),\label{eqn:Vdefinition}\\
\text{with} \quad 
  \Phi(k, k')  :=  \left( \prod_{i=k}^{k+k'-1} \tau(i)\right)^{-1}. \label{eqn:phi_definition}
\end{align}
This function satisfies the following inequalities
\begin{align}\label{eqn:V_bound}
\begin{aligned}
\dm\left(\xi,x_\ast\right) \leq V(k,  \xi) &\leq c_0 \dm\left(\xi,x_\ast\right)\\ 
V(k+1, f_k(\xi))&\leq \tau(k)V(k,  \xi) \end{aligned}\qquad\forall  k\geq 0, \quad \forall \xi \in \mathbb{R}^d. \end{align}
\end{theorem}
\begin{proof}First, we establish a lower bound for $V(k,\xi)$. Since $\Phi(k,0)=1$, we have
\begin{align*}
V(k,\xi) = \sup_{k' \geq 0}\dm\left(\phi(k',k, \xi),x_\ast\right) \Phi(k,k') \geq&~ \dm\left( \phi(0,k, \xi),x_\ast\right) \Phi(k,0)%\\&=\dm\left(\xi,x_\ast\right) \rho(0, k)^{-1}
= \dm\left(\xi,x_\ast\right).
\end{align*}
For the upper bound, from \eqref{eqn:stability}, we can write,
\begin{align*}
V(k, \xi) = \sup_{k' \geq 0}\dm\left(\phi(k',k, \xi),x_\ast\right) \Phi(k, k')&\leq \sup_{k' \geq 0}  c_0\dm\left(\xi,x_\ast\right)   \left( \prod_{i=k}^{k+k'-1} \tau(i)\right) \Phi(k, k')\\& \leq  \sup_{k' \geq 0}  c_0\dm\left(\xi,x_\ast\right)=c_0\dm\left(\xi,x_\ast\right).
\end{align*}
Next, we proceed by analyzing the behavior of the Lyapunov function for the subsequent time step, 
\begin{align*}
    V(k+1, \phi(1,k, \xi))&=\sup_{k' \geq 0}~\dm\left( \phi(k', \phi(1,k, \xi)) ,x_\ast\right) \Phi(k+1, k')\\&=\sup_{k' \geq 0}~\dm\left( \phi(k'+1,k, \xi),x_\ast\right)\Phi(k+1, k')\\
    &=\sup_{k' \geq 0}~\dm\left(\phi(k'+1,k, \xi),x_\ast\right) \Phi(k, k'+1)\tau(k)\\&\leq\tau(k)\sup_{k' \geq 1}~\dm\left(\phi(k',k, \xi),x_\ast\right) \Phi(k, k')\\
     &\leq \tau(k)~V(k,\xi),
\end{align*}
where we used the semi-group property of the flow map $\phi$ in the first step, the definition of $\Phi$ in the second step, and the monotonicity property of the supremum in the last step. This leads to the second line of the inequalities in \eqref{eqn:V_bound}.
\end{proof}

Theorem~\ref{thm:converselyapunov} thus provides upper and lower bounds on the corresponding Lyapunov function and establishes its rate-preserving property. Next, we show that the Lyapunov function is Lipschitz continuous on a compact subset of $\mathbb{R}^d$.  

\begin{proposition}\label{prop:L_V} %{(Lipschitz Continuity of the Lyapunov Function)}
Let Assumption~\ref{ass:uni_Lipschitz_f} hold and let $\mathcal{S}\subset \mathbb{R}^d$ be a compact set. Then there exists a Lipschitz constant $L_V$, such that the Lyapunov function defined in \eqref{eqn:Vdefinition} satisfies
\begin{equation*}
    |V(k,\xi_1)-V(k,\xi_2)|\leq L_V |\xi_1-\xi_2|, \quad \forall \xi_1,\xi_2\in \mathcal{S}, \quad\forall k\geq 0.
\end{equation*}
\end{proposition}
\begin{proof}  

From \eqref{eqn:ass_2} in Assumption~\ref{ass:convergence} we conclude that
\begin{equation*}
    \dm\left(\phi(k',k, \xi), x_\ast\right) \Phi(k,k') \leq \dm\left(\xi,x_\ast\right)\leq V(k,\xi)
\end{equation*}
for all $k'\geq K$, where $\Phi$ is defined in \eqref{eqn:phi_definition}. This implies that the supremum in $V(k,\xi)$ is attained at some finite index, which is bounded by $K$. As a result,
\begin{align}
V(k,\xi) = \max_{0 \leq k' \leq K} \dm\left(\phi(k',k, \xi), x_\ast\right) \Phi(k,k'), \quad \forall \xi \in \mathcal{S}, \quad \forall k\geq0.\end{align}
Now, for any $\xi_1, \xi_2 \in \mathcal{S}$, let
\begin{align*}
k^1 := \underset{{0 \leq k' \leq K}}{\operatorname{argmax}} ~\dm\left(\phi(k',k, \xi_1), x_\ast\right) \Phi(k,k'), \quad
k^2 := \underset{{0 \leq k' \leq K}}{\operatorname{argmax}} ~\dm\left(\phi(k',k, \xi_2), x_\ast\right) \Phi(k,k').
\end{align*}
Without loss of generality, assume $V(k,\xi_1) \geq V(k,\xi_2)$, then,
\begin{align*}
\left|V(k,\xi_1) - V(k,\xi_2)\right|&=V(k,\xi_1) - V(k,\xi_2)\\ &=  \dm\left(\phi(k^1, k,\xi_1), x_\ast\right) \Phi(k,k^1) - \dm\left(\phi(k^2,k, \xi_2), x_\ast\right) \Phi(k,k^2)\\ &\leq \dm\left(\phi(k^1, k,\xi_1), x_\ast\right) \Phi(k,k^1)  - \dm\left(\phi(k^1, k,\xi_2), x_\ast\right) \Phi(k,k^1)  \\&= \left( \dm\left(\phi(k^1,k, \xi_1), x_\ast\right) - \dm\left(\phi(k^1,k, \xi_2), x_\ast\right)\right)\Phi(k,k^1)\\
&\leq  \left( \dm\left(\phi(k^1,k, \xi_1), x_\ast\right) - \dm\left(\phi(k^1,k, \xi_2), x_\ast\right)\right)\Phi(0,k^1).
\end{align*}
In the third step, we used the fact that $$\dm\left(\phi(k^1,k, \xi_2), x_\ast\right)\Phi(k,k^1)\leq\dm\left(\phi(k^2,k,\xi_2), x_\ast\right) \Phi(k,k^2),$$ by definitions of $k^1$ and $k^2$. Then, the last step follows from the definition of $\Phi(k,k')$ in \eqref{eqn:phi_definition}.

Since the distance function is assumed to be norm-bounded \eqref{eqn:metric_norm_bound}, we have,
\begin{align*}
\left|\dm\left(\phi(k^1,k, \xi_1), x_\ast\right)  - \dm\left(\phi(k^1,k, \xi_2), x_\ast\right) \right| \leq L_\dm~ L_f^{k^1} |\xi_1 - \xi_2|.
\end{align*}
Multiplying both sides by $\Phi(0,k^1)$ gives,
\begin{align*}
\left|\dm\left(\phi(k^1,k,\xi_1), x_\ast\right) \Phi(0,k^1) - \dm\left(\phi(k^1,k, \xi_2), x_\ast\right) \Phi(0,k^1)\right| \leq L_\dm~ L_f^{k^1} \Phi(0,k^1) |\xi_1 - \xi_2|.
\end{align*}
By definition \eqref{eqn:phi_definition}, we have $\tau(i)\leq\tau(i+1)$ for all $i\geq0$, which implies $\Phi(0,k^1)\leq \tau(0)^{-k^1}$. Therefore, the following bound holds,
\begin{align*}
    |V(k,\xi_1) - V(k,\xi_2)| &\leq L_\dm~ L_f^{k^1} \tau(0)^{-k^1} |\xi_1 - \xi_2|\leq L_\dm~ L_f^{K} \tau(0)^{-K} |\xi_1 - \xi_2|.
\end{align*}for all $k>0$, where we used $k^1\leq K$ in the last step. Hence, we can set $L_V:=L_\dm~ L_f^{K} \tau(0)^{-K}$, which completes the proof.
\end{proof}
 
Once we determine the finite Lipschitz constant of the Lyapunov function, we can relate the disturbed system to the nominal system through the Lyapunov function, as established in Lemma~\ref{lemma:V_to_Vpert}.

\begin{lemma}\label{lemma:V_to_Vpert} Let Assumption~\ref{ass:bounded_effect_g} hold 
and let $\bar{V}:\mathbb{N} \times \mathbb{R}^d \rightarrow \mathbb{R}$ be any function that is $L_{V}-$Lipschitz in the second argument and satisfies the following condition,
\begin{align}
           \bar V(k+1,f_k(\xi))&\leq \tau(k) \bar V(k,\xi), \quad \forall \xi \in \mathbb{R}^d. \label{eqn:Vdecrease_inlemma}
\end{align}
Then, for the states of the disturbed system, the following inequality holds:
\begin{align}
           \bar V(k+1,g_k(\xi,e))&\leq \tau(k)\bar V(k,\xi) + L_V L_{\mathrm{e}k} |e|, \quad \forall \xi \in \mathbb{R}^d, ~\forall e \in \Omega,\label{eqn:Vpertdecrease_inlemma}
\end{align} 
where $|e|$ is the disturbance bound and $L_{\mathrm{e}k}$ is defined in Assumption~\ref{ass:bounded_effect_g}.   
\end{lemma}

\begin{proof}
We start with a Taylor series expansion of the Lyapunov function $V(k+1,g_k(z,e))$ around $e=0$,
\begin{align*}
   \bar V(k+1,g_k(z,e))=\bar V(k+1,g_k(z,0))+&\left.\frac{\partial \bar V(k+1,g_k(z,\bar{e}))}{\partial e}\right\rvert_{\bar{e}=\bar{\varepsilon}}e,
\end{align*}where $\bar{\varepsilon}$ is an intermediate point between $e$ and the origin. For the first part of the expansion, we evaluate $g_k(z,0)$:
\begin{align*}
  \bar  V(k+1,g_k(z,0))=\bar V(k+1,f_k(z)).
\end{align*}
Next, we analyze the derivative of the Lyapunov function with respect to $e$:
\begin{align*}
   \frac{\partial \bar V(k+1,g_k(z,e))}{\partial e} =\left.\frac{\partial\bar V(k+1,\bar{x})}{\partial x}\right\rvert_{\bar{x}=g_k(z,e)}\frac{\partial g_k(z,e)}{\partial e}.
\end{align*}
Substituting this back into the Taylor expansion yields,
\begin{align*}
  \bar  V(k+1,g_k(z,e))&=\bar V(k+1,f_k(z))+\left.\left.\frac{\partial \bar V(k+1,\bar{x})}{\partial x} \right\rvert_{\bar{x}=g_k(z,e)}\frac{\partial g_k(z,\bar{e})}{\partial e} \right\rvert_{\bar{e}=\bar{\varepsilon}}e.
\end{align*}
Since $g_k(x,e)$ satisfies Assumption~\ref{ass:bounded_effect_g}, its gradient is bounded by $L_{\mathrm{e}k}$. Additionally, $\bar V$ is $L_V-$Lipschitz with respect to its second argument. Thus, we arrive at
\begin{align*}
    \bar V(k+1,g_k(z,e))\leq\bar V(k+1,f_k(z))+L_V L_{\mathrm{e}k} |e|\leq \tau(k)\bar V(k,z)+L_V L_{\mathrm{e}k} |e|,
\end{align*} in view of \eqref{eqn:Vdecrease_inlemma}.
This concludes the proof.
\end{proof}

With these foundational results, we are now prepared to prove  Theorem~\ref{thm:mainthm}.

\begin{proof}
We prove Theorem~\ref{thm:mainthm} based on Theorem~\ref{thm:converselyapunov}, Proposition~\ref{prop:L_V}, Lemma~\ref{lemma:V_to_Vpert},  assuming that the Assumptions~\ref{ass:bounded_effect_g} and~\ref{ass:convergence} hold.
By Theorem~\ref{thm:converselyapunov}, there exists the Lyapunov function $V(k, \xi)$ that satisfies
\begin{align*}
V\left(k+1, f_k(\xi)\right) \leq \tau(k)~ V\left(k, \xi\right), \quad \forall\xi\in \mathbb{R}^d.
\end{align*}
From Proposition~\ref{prop:L_V}, we know that the Lyapunov function $V(k, \xi)$ is Lipschitz continuous with a finite Lipschitz constant $L_V$ on a compact set.

We then apply Lemma~\ref{lemma:V_to_Vpert}, which describes the behavior of $V$ under disturbances.
For the states of the disturbed system $z_k$, Lemma~\ref{lemma:V_to_Vpert} ensures,
\begin{align*}
V\left(k+1, z_{k+1}\right) \leq \tau(k) V\left(k, z_k\right)+L_V L_{\mathrm{e}k} |e_k|,
\end{align*}
where $L_V$ and $L_{\mathrm{e}k}$ are constants from Proposition~\ref{prop:L_V} and Assumption~\ref{ass:bounded_effect_g}, respectively.
By unrolling this recursion over a finite horizon $N>1$, we obtain,
\begin{align*}
V\left(N, z_{N}\right) \leq V\left(0, z_0\right)\left(\prod_{i=0}^{N-1} \tau(i)\right) +\sum_{j=0}^{N-1}L_V  \left(\prod_{i=j+1}^{N-1} \tau(i)\right) L_{\mathrm{e}j} |e_j|,
\end{align*} for $N\geq 1$.
Combining the results, we observe that the Lyapunov function for the disturbed system is controlled by the nominal system's convergence along with an additive term due to the disturbance. 

Incorporating the bounds for the Lyapunov function from Theorem~\ref{thm:converselyapunov}, we get
\begin{align*}
    \dm\left(z_k,x_\ast\right) \leq c_0 \left(\prod_{i=0}^{k-1} \tau(i)\right)\dm\left(z_0,x_\ast\right) +L_V   \sum_{j=0}^{k-1}\left(\prod_{i=j+1}^{k-1} \tau(i)\right)L_{\mathrm{e}j}|e_{j}| ,
\end{align*} for all $k\geq1$.
This inequality establishes the relationship between the system's state, the convergence rate, and the disturbances, thus proving Theorem~\ref{thm:mainthm}.
\end{proof}

\section{Applications to Learning Algorithms}\label{sec:applications}

Specific examples in the following sections highlight how this methodology can be used to analyze various learning algorithms, demonstrating its relevance in optimization, distributed systems, and privacy-preserving learning.

In each example, we first verify whether the learning algorithm satisfies the assumptions outlined in Assumption~\ref{ass:uni_Lipschitz_f} and Assumption~\ref{ass:convergence}. Next, we identify the disturbances affecting the learning algorithm. Understanding the nature and limits of these disturbances is essential for accurately quantifying their impact on stability. According to Assumption~\ref{ass:bounded_effect_g}, the dynamics of the perturbed system is Lipschitz continuous with the Lipschitz constant $L_{\mathrm{e}k}$. 
By substituting the value $c_0$ combined with the functions $L_{\mathrm{e}k}$ and $\tau(k)$, Theorem~\ref{thm:mainthm} delivers explicit expressions that describe the impact of disturbances on the algorithm's convergence up to a constant $L_V$ (Proposition~\ref{prop:L_V}). In the subsequent sections, we present specific examples demonstrating how our derived bounds apply to algorithms in different domains.

\subsection{Example: Distributed Optimization under Communication Constraints}\label{sec:opt}
Unlike traditional distributed (federated) learning algorithms, which rely on periodic updates across a randomly selected subset of agents \cite{Asad_2023}, event-based methods trigger communication only when necessary \cite{Er_2024, Cummins_2025, Zhang_2024}, significantly reducing overhead. 

Event-based communication strategies have been widely studied in networked systems for their ability to minimize unnecessary transmissions while maintaining system performance \cite{solowjow2020event, Singh_2023}. These methods monitor state changes and trigger updates based on predefined criteria, such as deviations that exceed a threshold \cite{Miskowicz_2006}. The communication threshold naturally introduces an error due to non-communicated updates in exchange for communication savings, which in turn impacts convergence. 

The systems-theoretic approach to the analysis of optimization algorithms allows us to model this error as a disturbance in an equivalent interconnected system that generates the same sequence as the event-based distributed optimization algorithm. For example, Er et al. \cite{Er_2024} model this error as an external bounded disturbance to the system and study the stability to achieve convergence bounds. Building on the methodology for proving linear convergence of Alternating Direction Method of Multipliers (ADMM) outlined in \cite{Nishihara_2015}, \cite{Er_2024} extends this result to the disturbed case.
The analysis highlights the inherent trade-offs between event thresholds, convergence accuracy, and communication efficiency. Larger thresholds reduce communication demands, but may slow convergence and increase suboptimality, while smaller thresholds ensure higher accuracy at the cost of more frequent transmissions.

We recover the earlier results from \cite{Er_2024} using Theorem~\ref{thm:mainthm}. Let us consider an objective function $\ell=\sum_{i=1}^N \ell^i$ that is $\gamma$-strongly convex and $\beta$-smooth, with a condition number $\kappa={\beta}/{\gamma}$. We analyze a relaxed ADMM variant with the relaxation parameter $\alpha$ and a step size determined by $h=\kappa^\epsilon\sqrt{\gamma\beta}$, with tuning parameter $\epsilon>0$. 

Nishihara et al. \cite{Nishihara_2015} establish a decay rate of ADMM as 
$\tau=1-{\alpha}/({2\kappa^{\epsilon+1/2}})$ using semidefinite programming.
Their approach generalizes prior ADMM analyses by reducing convergence proofs to verifying the feasibility of a linear matrix inequality. This procedure leads to the following bound,
\begin{align*}
\left|\theta_k-\theta_\ast\right| \leq {\tau}^{k} c_0\left|\theta_0-\theta_\ast\right|,
\end{align*}
for the nominal (undisturbed) dynamics, and for a constant $c_0$. 
We then introduce a marginally slower rate 
$\tilde{\tau}=1-{\alpha/}({4\kappa^{\epsilon+1/2}})$ (still with the same scaling in $\kappa$) to ensure \eqref{eqn:ass_2} is satisfied.

The event-based communication scheme that we will analyze employs a constant threshold that triggers communication if the change in the local variable is large compared to the last communication event. 
The disturbance in this setting is the difference between the iterate that would be sent with continuous communication and the iterate sent under the event-based rule. In other words, it measures how much the local model has drifted from the value that was used to form the current global model. This deviation is bounded because the event-based communication scheme enforces that each agent communicate whenever its local state deviates too far from the last transmitted value.
Therefore, the disturbance satisfies $\left|e_k\right| \leq \Delta$. 
Substituting these into Theorem~\ref{thm:mainthm}, we obtain the bound for the disturbed system,
\begin{align*}
\left|\theta_k-\theta_\ast\right| \leq \tilde{\tau}^{k} c_0\left|\theta_0-\theta_\ast\right| +\frac{L_V L_e \Delta}{1-\tilde{\tau}}.
\end{align*}
For large $k$, the transient term $\tilde{\tau}^{k} c_0|\theta_0-\theta_\ast|$ vanishes as $\tilde{\tau}^k \rightarrow 0$, and the steady-state bound becomes $|\theta_k-\theta_\ast| \approx {L_V L_e \Delta}\kappa^{\epsilon+1/2}/{\alpha}$. This result quantifies the trade-off between the disturbance magnitude $\Delta$ and the decay rate $\tilde{\tau}$, which depends on the condition number $\kappa$.

It is important to note that the bound in prior work \cite{Er_2024} depends on a handcrafted quadratic Lyapunov function and corresponding linear matrix inequality. Theorem~\ref{thm:mainthm} yields a tighter bound of $\mathcal{O}({\kappa^{\frac{1}{2}+\epsilon}\Delta}/{\alpha})$ compared to the {$\mathcal{O}({\kappa^{1+\epsilon}\Delta}/({\alpha\min\{\alpha,2-\alpha\}}))$} bound reported in \cite{Er_2024}. This improvement suggests that a different Lyapunov certificate can lead to sharper convergence guarantees. In fact, by modifying the tradeoff parameters, we obtain a bound of $\mathcal{O}(\kappa^{\frac{1}{2}+\epsilon}\Delta)$ which aligns with the results derived using Theorem~\ref{thm:mainthm}.

\subsection{Example: Generalization and Algorithmic Performance}\label{sec:gen_stb}
Prior work \cite{Hardt_2016} has demonstrated that algorithmic stability and generalization performance are intrinsically linked. Recent analyses \cite{Kozachkov_2023} formalize and strengthen these insights using contraction theory. {In particular, using contraction analysis, Kozachkov et al. \cite{Kozachkov_2023} showed that Riemannian contraction guarantees generalization in supervised learning.} We extend this analysis by considering stability conditions beyond contraction and show that meaningful generalization bounds still emerge. Here, we demonstrate how Theorem~\ref{thm:mainthm} reproduces well-known results from \cite{Hardt_2016} regarding the algorithmic stability of stochastic gradient descent on both strongly convex and convex losses.

We consider parameter updates for two adjacent datasets $D$ and $D'$, differing by only one element. Our goal is to bound the effect of the one data point on the resulting model, i.e., $ |\theta^{D'}_{k}-\theta^{D}_{k}|$. We use the Euclidean norm as our metric,
\begin{align*}
    \dm\left( \theta^{D'}_{k},\theta^{D}_\ast \right) = \left|\theta^{D'}_{k}-\theta^{D}_\ast\right|.
\end{align*}
For a $\gamma-$strongly convex, $\beta-$smooth, $L_\ell-$Lipschitz loss $\ell$, and  step size of $h=2/{(\beta+\gamma)}$, the following convergence rate holds \cite[Theorem 2.1.15]{Nesterov_2018},
\begin{align}
    \left|\theta^{D}_{k}-\theta^{D}_\ast\right|&\leq \left(\frac{\beta-\gamma}{\beta+\gamma}\right)^k\left|\theta^{D}_{0}-\theta^{D}_\ast\right|. \label{eqn:genD}\end{align} 

Theorem~\ref{thm:mainthm} implies that for the disturbed parameter sequence associated with $D'$, the following holds,
    \begin{align}
    \left|\theta^{D'}_{k}-\theta^{D}_\ast\right|&\leq \tilde{\tau}^k c'_0\left|\theta^{D'}_{0}-\theta^{D}_\ast\right|+ L_V L_e\sum_{j=0}^{k} |e_j|\tilde{\tau}^{k-j-1},\label{eqn:genD'}
\end{align} where the convergence rate $\tilde{\tau}={(\beta-\delta\gamma)}/{(\beta+\gamma)}$ with $0<\delta\leq 1$ ensures that \eqref{eqn:ass_2} is satisfied, $L_e$ is equal to the chosen step size $h= {2}/{(\beta+\gamma)}$ and the effect of one data point (i.e., disturbance) is bounded by the gradient bound scaled by the number of data points, $|e_j|\leq {L_\ell}/{n}$.
By adding \eqref{eqn:genD} and \eqref{eqn:genD'}, and applying the triangle inequality, we obtain
\begin{align*}
        \left|\theta^{D'}_{k}-\theta^{D}_{k}\right|\leq \tilde{\tau}^kC+ L_V \frac{h L_\ell}{n}\sum_{j=1}^{k-1}\tilde{\tau}^{k-j-1},
\end{align*} where $C$ collects the constants from \eqref{eqn:genD} and \eqref{eqn:genD'}. 
Therefore, the stability bound $\epsilon_{\text{stab}}:=\lim_{k\rightarrow\infty}|\ell(\theta^{D'}_{k})-\ell(\theta^{D}_{k})|$ satisfies 
\begin{align*}
    \epsilon_\text{stab}& \leq  \lim_{k\rightarrow\infty} L_V \frac{h L_\ell^2}{n}\sum_{j=1}^{k-1} \tilde{\tau}^{k-j-1}.\end{align*}
Computing the sum and substituting $h={2}/{(\beta+\gamma)}$ and $\tilde{\tau}$, we have   
\begin{align*}
       \epsilon_\text{stab}  \leq \lim_{k\rightarrow\infty} L_V \frac{2L_\ell^2}{(\beta+\gamma) n} \frac{1-\tilde{\tau}^{k}}{1-\tilde{\tau}}\leq\frac{2L_V L_\ell^2}{\gamma (\delta+1) n}%\leq \frac{L_V L_\ell^2 }{\beta n(1-\tau)}\overset{(\tau=1-\frac{\gamma}{\beta})}{=}\frac{L_V L_\ell^2 }{\beta n\frac{\gamma}{\beta}}
       \leq\frac{2L_V L_\ell^2}{\gamma n}.
\end{align*}
This aligns with \cite[Theorem 3.9]{Hardt_2016}, which states $\epsilon_\text{stab}\leq{2L_\ell^2}/{(\gamma n)}$.

For a convex, $\beta$-smooth loss, we characterize stability by adapting  Nesterov’s last iterate convergence for convex (but not strongly convex) objectives  with a time-varying step size  $h_k\leq2/\beta$ \cite[Theorem 2.1.14]{Nesterov_2018}. In this setting, we have
\begin{align*}
    \left|\ell(\theta^{D}_{k})-\ell(\theta^{D}_\ast)\right|&\leq \frac{\left|\theta^{D}_{0}-\theta^{D}_\ast\right|^2}{\frac{2}{\beta}+\sum_{l=0}^kh_l\left(1-\frac{\beta}{2}h_l\right)}.
\end{align*}

We now use Theorem~\ref{thm:mainthm} to state the convergence of the parameter sequence associated with the adjacent dataset,
\begin{align*}
    \left|\ell(\theta^{D'}_{k})-\ell(\theta^{D}_\ast)\right|&\leq c_0\left(\prod_{i=0}^{k-1} \tilde{\tau}(i)\right)\left|\theta^{D}_{0}-\theta^{D}_\ast\right|^2+ L_V   \sum_{j=0}^{k-1}\left(\prod_{i=j+1}^{k-1} \tilde{\tau}(i)\right)L_{\mathrm{e}j}|e_{j}|, 
\end{align*}
where $\tilde{\tau}(j)=\left(1-{1}/{(j+2)}\right)^c$ for some small $c>0$. 
Similar to the first case, by triangle inequality and plugging in $L_{\mathrm{e}k}=L_\ell h_k$ and $|e_j|\leq L_\ell/n$, we obtain
\begin{align*}
    \epsilon_\text{stab}\leq \lim_{k\rightarrow\infty} L_V \frac{L_\ell^2}{n} \sum_{j=0}^{k-1}\left( h_{j}\prod_{i=j+1}^{k-1} \tilde{\tau}(i)\right).
\end{align*}

Applying Hölder’s inequality with $(\infty,1)$ (Remark~\ref{remark:holder}), we deduce that
\begin{align*}
    \epsilon_\text{stab}\leq \lim_{k\rightarrow\infty} L_V \frac{L_\ell^2}{n}\max_{0\leq l \leq k}\left|\prod_{i=l+1}^{k-1} \tilde{\tau}(i)\right|\sum_{j=0}^{k-1} h_{j}{\leq} \lim_{k\rightarrow\infty} \frac{L_V L_\ell^2}{n}\sum_{j=0}^{k-1} h_{j}.
\end{align*}

This expression matches the bound in \cite[Theorem 3.8]{Hardt_2016} and \cite[Section 4.2.1]{Kozachkov_2023}, where the bound is represented by $\tfrac{2L_\ell^2}{n}\sum_{l=1}^kh_l$ with step size satisfying $h_l\leq {2}/{\beta}$.

% Conclusion
These results show that our systems-theoretic framework naturally recovers and extends the classical generalization bounds derived from algorithmic stability. 
In particular, Theorem~\ref{thm:mainthm} not only reproduces known results for both strongly convex and convex settings but also offers a structural interpretation of how algorithm properties such as smoothness and step size determine generalization performance. This highlights the importance of Lyapunov-based reasoning in understanding stability.

\subsection{Example: Privacy-Preserving Learning Mechanisms}\label{sec:dp}
A standard approach to ensuring differential privacy in iterative optimization is to perturb each gradient update with Gaussian noise \cite{andrew_2021_dp, chaudhuri_2011_dp}. This results in noisy gradient descent,
whose utility bounds are well studied in the literature \cite{Shalev_2009, Shamir_Zhang_2013, Bassily_2014, Bassily_2019, Altschuler_Talwar_2023}. In this section, we reformulate the bounds with our systems-theoretic approach. By Theorem~\ref{thm:zero_mean_stochastic}, the effect of additive Gaussian noise on convergence can be quantified directly through stability properties of the underlying dynamics.

We start with the noisy gradient descent update rule,
\begin{align*}
  \theta_{k+1}=\theta_{k}-h\left(\nabla \ell(\theta_{k})+n_{k}\right), \quad n_{k} \overset{\text{i.i.d.}}{\sim} \mathcal{N}(0,\sigma^2_n\mathrm{I}_d/d), 
\end{align*}
where $h$ denotes the step size, $\nabla \ell(\theta_{k})$ represents the gradient evaluated at the current parameter $\theta_{k}$ (on the fixed dataset ${D}$), and $n_{k}$ is additive Gaussian noise with variance $\sigma^2_n/d$, where $d$ denotes the dimension. 
In practice, privacy-preserving variants of stochastic gradient descent adopt exactly this mechanism, replacing $\nabla \ell(\theta_{k})$ by a stochastic gradient and injecting Gaussian noise \cite{Bassily_2014}. Our analysis focuses on the core effect of noise injection; the same reasoning applies when stochastic gradients are also present.

We assume the objective is $\gamma$-strongly convex and $\beta$-smooth, the chosen constant step size is $h \leq {2}/({\beta + \gamma})$. In this setting, classical analysis yields the deterministic linear decay \cite[Theorem 2.1.15]{Nesterov_2018}, with $\tau(j)=\tau\in(0,1]$ for all $j\geq 0$, where $\tau = 1 - \tfrac{2h\gamma\beta}{\beta + \gamma}$,
\begin{align*}
\ell(\theta_{k})-\ell(\theta_\ast)\leq \left(1-\frac{2h\gamma\beta}{\beta+\gamma}\right)^k\frac{\beta}{\gamma}(\ell(\theta_{0})-\ell(\theta_\ast)).
\end{align*} 

Applying Theorem~\ref{thm:zero_mean_stochastic} to the noisy dynamics with  the Lipschitz constant of the disturbance map as $L_\mathrm{e}=L_\ell h$, 
we obtain the expected error bound 
\begin{align*}
\mathbb{E}\left\{  \ell(\theta^{\text{priv}}_{k})-\ell(\theta_\ast)\right\} \leq\frac{\beta}{\gamma}\tau^k(\ell(\theta_{0})-\ell(\theta_\ast))+ \frac{1}{2}L_HL_\ell^2\sigma_n^2h^2\frac{1}{1-\tau}.
\end{align*} 

Since $\log(\tau) \leq -\tfrac{2h\gamma\beta}{\beta+\gamma}$, choosing
$h = \tfrac{\beta+\gamma}{2\gamma\beta}\tfrac{\log N}{N}$ for horizon $N$ ensures
\begin{align*}
   \frac{1}{2}\frac{L_HL_\ell^2\sigma_n^2h^2}{1-\tau}\leq \frac{1}{4}{L_HL_\ell^2\sigma_n^2h}\frac{(\beta+\gamma)}{\gamma\beta} \leq \frac{1}{8}  L_HL_\ell^2\sigma_n^2\frac{(\beta+\gamma)^2}{\gamma^2\beta^2}\frac{\log N}{N},
\end{align*}
so that the expected error becomes
    $\mathbb{E}\{ \ell(\theta^{\text{priv}}_{N})-\ell(\theta_\ast)\}=\mathcal{O}\left(\sigma_n^2 \beta^2 {\log N}/({\gamma^2 N)}\right)$.

We further showcase the flexibility of our method by deriving bounds for accelerated gradient descent, which, for example, takes the following update rule,
\begin{align*}
\theta_{k+1}=\theta_k+h p_{k+1}, \quad p_{k+1}=(1-2\bar{d} h)p_k-h (\nabla \ell(\theta_k+\bar{\beta} p_k) +n_k)/\beta,
\end{align*}
where $\bar{d}=1/(1+\sqrt{\beta/\gamma})$ and $\bar{\beta}=1-2\bar{d}$ are damping parameters and $h \leq 1$ is the step size. The analysis in \cite[App.~A6]{muehlebach_icml} proves the following nominal convergence rate
\begin{equation*}
    \ell(\theta_k+c_1p_k)-\ell(\theta_\ast) \leq \frac{\beta}{\gamma} (1-\bar{d}h)^k (\ell(\theta_0+c_1p_0)-\ell(\theta_\ast)),
\end{equation*}
for any $h\in (0,1]$, where $c_1=\bar{\beta}/(1-2\bar{d}h)-h$. Applying Theorem~\ref{thm:zero_mean_stochastic} to the noisy dynamics with the Lipschitz constant of the disturbance map as $L_\mathrm{e}=L_\ell h/\beta$, we obtain
\begin{align*}
\mathbb{E}\left\{  \ell(\theta^{\text{priv}}_{k}+c_1 p_k^\text{priv})-\ell(\theta_\ast)\right\} \leq\frac{\beta}{\gamma} (1-\bar{d}h)^k(\ell(\theta_{0}+c_1p_0)-\ell(\theta_\ast))+ \frac{1}{2}L_HL_\ell^2\sigma_n^2 \frac{h}{\beta^2 \bar{d}}.
\end{align*}
Applying a similar argument as before and choosing the step size as $h=\text{log}(N)/(N\bar{d})$ results in a bound of $\mathcal{O}\left(\sigma_n^2 \beta {\log N}/({\gamma N)}\right)$. We note that, compared to gradient descent, the bound improves by a factor of $\beta/\gamma$, which can be very substantial.

Our results highlight how system dynamics and noise sensitivity jointly shape the trade-off between convergence rate and utility loss due to additional noise. 

\section{Conclusions}
\label{sec:conclusions}
In this article, we have developed a methodology for stability analysis of learning algorithms through a systems-theoretic view. Through our main result in Section~\ref{sec:main}, we lay out a blueprint for the analysis of perturbed algorithms based on the characteristics of their unperturbed counterparts. Examples we presented in Section~\ref{sec:applications} show three different cases of perturbation: disturbance that naturally arises due to system constraints (Section~\ref{sec:opt}), modeled disturbance to analyze the sensitivity to one data point (Section~\ref{sec:gen_stb}), and deliberately injected disturbance to achieve differential privacy (Section~\ref{sec:dp}). These insights advance our understanding of stability and robustness in learning algorithms and help us both design and model disturbance mechanisms to ensure desirable algorithmic properties.

\section*{Acknowledgments}
Guner Dilsad ER and Michael Muehlebach thank the German Research Foundation (grant number 456587626 of the Emmy Noether program) and the International Max Planck Research School for Intelligent Systems (IMPRS-IS) for their support.

\bibliographystyle{siamplain}
\bibliography{ref}

@article{Nishihara_2015, title={A General Analysis of the Convergence of {ADMM}}, journal={Proceedings of the International Conference on Machine Learning}, author={Nishihara, Robert and Lessard, Laurent and Recht, Benjamin and Packard, Andrew and Jordan, Michael I.}, year={2015}, volume={37}, pages={343-352}}

@article{Lessard_2016, title={Analysis and Design of Optimization Algorithms via Integral Quadratic Constraints}, volume={26}, number={1}, journal={SIAM Journal on Optimization}, author={Lessard, Laurent and Recht, Benjamin and Packard, Andrew}, year={2016}, pages={57-95}}

@article{muehlebach_icml, title = {A Dynamical Systems Perspective on {Nesterov} Acceleration}, author = {Muehlebach, Michael and Jordan, Michael I}, year ={2019}, journal={Proceedings of the International Conference on Machine Learning}, volume={97}, pages={4656-4662}}

@article{jordan_variational_2016, title = {A variational perspective on accelerated methods in optimization}, author = {Wibisono, Andre and Wilson, Ashia C. and Jordan, Michael I}, year ={2016}, volume={113}, number={47}, pages={E7351-E7358}, journal={Proceedings of the National Academy of Sciences}}

@article{su_differential_2016, title = {A Differential Equation for Modeling {Nesterov}’s Accelerated Gradient Method: {T}heory and Insights}, author = {Su, Weijie and Boyd, Stephen and Candes, Emmanuel J}, journal={Journal of Machine Learning Research},  volume={17}, year={2016}, pages={1-43}, number={153}}

@article{Tong_2023, title = {A Dynamical Systems Perspective on Discrete Optimization}, author = {Tong, Guanchun and Muehlebach, Michael}, journal = {Proceedings of Machine Learning Research}, volume={211}, pages={1-14}, year = {2023}}

@article{Muehlebach_2020, title = {Continuous-time Lower Bounds for Gradient-based Algorithms}, author = {Muehlebach, Michael and Jordan,  Michael I.}, journal = {Proceedings of the International Conference on Machine Learning}, year = {2020}, pages={7088-7096}, volume={119}}

@article{JMLR:v23:21-0798,
  author  = {Michael Muehlebach and Michael I. Jordan},
  title   = {On Constraints in First-Order Optimization: {A} View from Non-Smooth Dynamical Systems},
  journal = {Journal of Machine Learning Research},
  year    = {2022},
  volume  = {23},
  number  = {256},
  pages   = {1-47}
}

@article{van2017fastest,
  title={The fastest known globally convergent first-order method for minimizing strongly convex functions},
  author={Van Scoy, Bryan and Freeman, Randy A and Lynch, Kevin M},
  journal={IEEE Control Systems Letters},
  volume={2},
  number={1},
  pages={49-54},
  year={2017}
}

@article{scherer2021convex,
  title={Convex synthesis of accelerated gradient algorithms},
  author={Scherer, Carsten and Ebenbauer, Christian},
  journal={SIAM Journal on Control and Optimization},
  volume={59},
  number={6},
  pages={4615-4645},
  year={2021}
}

@article{jakob2025,
    author={Fabian Jakob and Andrea Ianelli},
    title={Online Convex Optimization and Integral Quadratic Constraints: {A} new approach to regret analysis},
    journal={arXiv:2503.23600},
    pages={1-7},
    year={2025}
}

@article{zhang2025,
  title = {Primal Methods for Variational Inequality Problems with Functional Constraints},
  journal = {Mathematical Programming},
  year = {2025},
  author = {Zhang, Liang and He, Niao and Muehlebach, Michael},
  pages={1-32}  
}

@article{Attouch,
	author={Hedy Attouch and Zaki Chbani and Juan Peypouquet and Patrick Redont},
	title={Fast convergence of inertial dynamics and algorithms with asymptotic vanishing viscosity},
	journal={Mathematical Programming},
	volume={168},
	number={1-2},
	pages={123-175},
	year={2018}
}

@article{Attouch1,
    author={Felipe Alvarez and Hedy Attouch},
    title={An Inertial Proximal Method for Maximal Monotone Operators via Discretization of a Nonlinear Oscillator with Damping},
    journal={Set-Valued and Variational Analysis},
    volume={9},
    year={2001},
    pages={3-11}
}

@article{Attouch2,
    author={Hedy Attouch and Paul-\'{E}mile Maing\'{e}},
    title={Asymptotic behavior of second-order dissipative evolution equations combining potential with non-potential effects},
    journal={ESAIM: Control, Optimisation and Calculus of Variations},
    year={2011},
    volume={17},
    number={3},
    pages={836-857}
}

@article{Attouch3,
    author={Hedy Attouch},
    title={Fast inertial proximal {ADMM} algorithms for convex structured optimization with linear constraint},
    journal={Minimax Theory and its Applications},
    volume={6},
    number={1},
    pages={1-24},
    year={2021}
}

@article{Attouch4,
    author={Hedy Attouch and Zaki Chbani and Jalal Fadili and Hassan Riahi},
    title={Fast convergence of dynamical {ADMM} algorithms for convex structured optimization with linear constraint},
    journal={Journal of Optimization Theory and Applications},
    volume={193},
    number={1-3},
    pages={704-736},
    year={2022}
}

@article{Mathprog,
  title={Accelerated first-order optimization under nonlinear constraints},
  author={Muehlebach, Michael and Jordan, Michael I},
  journal={Mathematical Programming},
  pages={1-46},
  year={2025},
  publisher={Springer}
}

@article{wibisono2018sampling,
  title={Sampling as optimization in the space of measures: {T}he {L}angevin dynamics as a composite optimization problem},
  author={Wibisono, Andre},
  journal={Proceedings of the Conference on Learning Theory},
  pages={2093-3027},
  year={2018},
  volume={75}
}

@article{guo2024provable,
  title={Provable benefit of annealed {L}angevin {M}onte {C}arlo for non-log-concave sampling},
  author={Guo, Wei and Tao, Molei and Chen, Yongxin}, journal={Proceedings of the International Conference on Learning Representations},
  year={2025},
  pages={1-23}
}

@article{song2020score,
  title={Score-based generative modeling through stochastic differential equations},
  author={Song, Yang and Sohl-Dickstein, Jascha and Kingma, Diederik P and Kumar, Abhishek and Ermon, Stefano and Poole, Ben},
  journal={Proceedings of the International Conference on Learning Representations},
  year={2021},
pages={1-36}
}

@article{berner2022optimal,
  title={An optimal control perspective on diffusion-based generative modeling},
  author={Berner, Julius and Richter, Lorenz and Ullrich, Karen},
  journal={arXiv:2211.01364},
  pages={1-42},
  year={2022}
}

@book{hahn1967stability,
  title={Stability of motion},
  author={Hahn, Wolfgang},
  year={1967},
  publisher={Springer}
}

@article{vempala2019rapid,
  title={Rapid convergence of the unadjusted {L}angevin algorithm: {I}soperimetry suffices},
  author={Vempala, Santosh and Wibisono, Andre},
  journal={Proceedings of the International Conference on Advances in Neural Information Processing Systems},
  volume={32},
  year={2019},
pages={8094-8106}
}

@article{schechtman2023orthogonal,
  title={Orthogonal Directions Constrained Gradient Method: from non-linear equality constraints to {Stiefel} manifold},
  author={Schechtman, Sholom and Tiapkin, Daniil and Muehlebach, Michael and Moulines, Eric},
  journal={Proceedings of the Conference on Learning Theory},
  pages={1228-1258},
  year={2023}
}

@article{Er_2024,
    author = {Guner Dilsad Er and Sebastian Trimpe and Michael Muehlebach},
    title = {Distributed Event-Based Learning via {ADMM}},
    journal={Proceedings of the International Conference on Machine Learning},
    year = {2025},
    pages={15384-15418}, volume={267}
}

@article{Hardt_2016, title={Train faster, generalize better: {S}tability of stochastic gradient descent}, author={Hardt, Moritz and Recht, Benjamin and Singer, Yoram}, year={2016}, journal={Proceedings of the International Conference on Machine Learning}, volume={48}, pages={1225-1234}}

@article{Kozachkov_2023, title={Generalization as Dynamical Robustness -- The Role of {R}iemannian Contraction in Supervised Learning},  author={Kozachkov, Leo and Wensing, Patrick M and Slotine, Jean-Jacques}, year={2023}, journal={Transactions on Machine Learning Research}}

@article{Wensing_Slotine_2020, title={Beyond Convexity -- Contraction and Global Convergence of Gradient Descent}, volume={15}, number={8}, journal={PLOS ONE}, author={Wensing, Patrick M. and Slotine, Jean-Jacques E.}, year={2020}, pages={e0236661} }

@book{absil2009optimization,
  author={Absil, P-A and Mahony, Robert and Sepulchre, Rodolphe},
  title={Optimization Algorithms on Matrix Manifolds},
  year={2009},
  publisher={Princeton University Press}
}

@book{Sastry_1999, 
title={Nonlinear Systems}, 
publisher={Springer}, author={Sastry, Shankar}, year={1999}
}

@book{Khalil_2002nonlinear,
  title={Nonlinear systems},
  author={Khalil, Hassan},
  year={2002},
  publisher={Prentice Hall}
}

@article{Jiang_1994, title={Small-gain theorem for {ISS} systems and applications}, volume={7},  number={2}, journal={Mathematics of Control, Signals, and Systems}, author={Jiang, Z. -P. and Teel, A. R. and Praly, L.}, year={1994}, pages={95–120}}

@article{Jiang_Wang_2002_conv, title={A converse {Lyapunov} theorem for discrete-time systems with disturbances}, volume={45}, number={1}, journal={Systems \& Control Letters}, author={Jiang, Zhong-Ping and Wang, Yuan}, year={2002}, pages={49–58} }

@article{Sontag_Wang_1996_new_char, title={New characterizations of input-to-state stability}, volume={41}, number={9}, journal={IEEE Transactions on Automatic Control}, author={Sontag, Eduardo and Yuan Wang}, year={1996},  pages={1283–1294} }

@inbook{Sontag_book_2008,
title={Input to State Stability: {B}asic Concepts and Results}, volume={1932},  booktitle={Nonlinear and Optimal Control Theory}, publisher={Springer}, author={Sontag, Eduardo},  year={2008}, pages={163–220}}

@article{Zhang_2024, title={Privacy-preserving distributed {ADMM} with event-triggered communication}, journal={IEEE Transactions on Neural Networks and Learning
Systems}, author={Zhang, Z. and Yang, S. and Xu, W. and Di, K.}, volume={35}, number={2}, pages={2835–2847}, year={2024}}

@article{solowjow2020event, title={Event-triggered learning}, author={Solowjow, Friedrich and Trimpe, Sebastian},  journal={Automatica},  volume={117},  pages={109009},  year={2020}}

@article{Singh_2023, title={{SPARQ-SGD}: {E}vent-Triggered and Compressed Communication in Decentralized Optimization}, volume={68}, number={2}, journal={IEEE Transactions on Automatic Control}, author={Singh, Navjot and Data, Deepesh and George, Jemin and Diggavi, Suhas}, year={2023}, pages={721-736} }

@article{Miskowicz_2006, title = {Send-{On}-{Delta} Concept: {A}n Event-Based Data Reporting Strategy},	volume = {6}, number = {1}, journal = {Sensors}, author = {Miskowicz, Marek}, year = {2006}, pages = {49-63},
}

@article{Asad_2023, title={Limitations and Future Aspects of Communication Costs in Federated Learning: {A} Survey}, volume={23}, number={17}, journal={Sensors}, author={Asad, Muhammad and Shaukat, Saima and Hu, Dou and Wang, Zekun and Javanmardi, Ehsan and Nakazato, Jin and Tsukada, Manabu}, year={2023}, pages={7358}}

@article{Shamir_Zhang_2013, title={Stochastic Gradient Descent for Non-smooth Optimization: {C}onvergence Results and Optimal Averaging Schemes},  journal={Proceedings of the International Conference on Machine Learning}, volume ={28}, author={Shamir, Ohad and Zhang, Tong}, year={2013}, number={1}, pages={71-79} }

@book{Nesterov_2018, title={Lectures on Convex Optimization},  publisher={Springer}, author={Nesterov, Yurii}, year={2018}}

@article{Cummins_2025,
author = {Cummins, Michael and Er, Guner Dilsad and Muehlebach, Michael},
title = {Controlling Participation in Federated Learning with Feedback},
journal = {Proceedings of the Learning for Dynamics and Control Conference},
pages={1-13},
volume = {283},   
year = {2025} }

@article{Dörfler_2024, title={Toward a Systems Theory of Algorithms}, volume={8}, journal={IEEE Control Systems Letters}, author={Dörfler, Florian and He, Zhiyu and Belgioioso, Giuseppe and Bolognani, Saverio and Lygeros, John and Muehlebach, Michael}, year={2024}, pages={1198–1210} }

@article{Sontag_Wang_1995, title={On characterizations of the input-to-state stability property}, volume={24}, number={5}, journal={Systems \& Control Letters}, author={Sontag, Eduardo and Wang, Yuan}, year={1995}, month=apr, pages={351–359} }

@book{Helmke_2012, author= {U. Helmke and J. B. Moore}, title={Optimization and Dynamical Systems}, publisher ={Springer}, year= {2012} }

@article{Brockett_1991, 
author= {R. W. Brockett}, title={Dynamical systems that sort lists, diagonalize matrices, and solve linear programming problems}, journal={Linear Algebra Applications}, 
volume = {146}, 
pages = {79–91}, 
year={1991}
}

@book{Stuart_1998, author= {A. Stuart and A. R. Humphries}, title={Dynamical Systems and Numerical Analysis}, publisher={Cambridge University Press}, year={1998}}

@article{Kashima_2007,
author = {Kenji Kashima and Yutaka Yamamoto},
title = {System theory for numerical analysis},
journal = {Automatica},
year = {2007},
volume = {43},
publisher = {Elsevier},
number = {7},
pages = {1156-1164}}

@article{Altschuler_Talwar_2023, title={Privacy of Noisy Stochastic Gradient Descent: {M}ore Iterations without More Privacy Loss}, 
journal = {Proceedings of the International Conference on Advances in Neural Information Processing Systems},
pages = {3788-3800}, author={Altschuler, Jason M. and Talwar, Kunal}, year={2022}, volume = {35}}

@article{Bassily_2019, title={Private Stochastic Convex Optimization with Optimal Rates}, journal={arXiv:1908.09970}, author={Bassily, Raef and Feldman, Vitaly and Talwar, Kunal and Thakurta, Abhradeep}, year={2019}, pages={1-18}}

@article{Bassily_2014, title={Differentially Private Empirical Risk Minimization: {E}fficient Algorithms and Tight Error Bounds},  journal={arXiv:1405.7085}, author={Bassily, Raef and Smith, Adam and Thakurta, Abhradeep}, year={2014}, pages={1-39} }

@article{Shalev_2009, title={Stochastic Convex Optimization}, author={Shalev-Shwartz, Shai and Shamir, Ohad and Srebro, Nathan and Sridharan, Karthik}, journal = {Proceedings of the Conference on Learning Theory}, volume = {2}, number ={4}, page={5}, year={2009} }

@article{Perdomo_2020,
  title = 	 {Performative Prediction},
  author =       {Perdomo, Juan and Zrnic, Tijana and Mendler-D{\"u}nner, Celestine and Hardt, Moritz},
  journal = 	 {Proceedings of the International Conference on Machine Learning},
  pages = 	 {7599-7609},
  year = 	 {2020},
  volume = 	 {119}}

@article{Stich_2018,
author = {Stich, Sebastian U. and Cordonnier, Jean-Baptiste and Jaggi, Martin},
title = {Sparsified {SGD} with memory},
year = {2018},
journal = {Proceedings of the International Conference on Advances in Neural Information Processing Systems},
pages = {4452–4463},
numpages = {12},
volume={31}
}

@article{Alistarh_2017,
 author = {Alistarh, Dan and Grubic, Demjan and Li, Jerry and Tomioka, Ryota and Vojnovic, Milan},
 journal = {Proceedings of the International Conference on Advances in Neural Information Processing Systems},
 editor = {I. Guyon and U. Von Luxburg and S. Bengio and H. Wallach and R. Fergus and S. Vishwanathan and R. Garnett},
 pages = {1709-1720},
 title = {{QSGD}: {C}ommunication-Efficient {SGD} via Gradient Quantization and Encoding},
 volume = {30},
 year = {2017}
}

@article{Zhiyu_2025,
  title={Decision-Dependent Stochastic Optimization: {The} Role of Distribution Dynamics},
  author={He, Zhiyu and Bolognani, Saverio and Dörfler, Florian and Muehlebach, Michael},
  journal={arXiv:2503.07324},
  pages={1-47},
  year={2025}
}

@article{Cao_2023,
author={Cao, Xuanyu and Başar, Tamer and Diggavi, Suhas and Eldar, Yonina C. and Letaief, Khaled B. and Poor, H. Vincent and Zhang, Junshan},
  journal={IEEE Journal on Selected Areas in Communications}, 
  title={Communication-Efficient Distributed Learning: An Overview}, 
  year={2023},
  volume={41},
  number={4},
  pages={851-873}}

@article{FedPAQ_2020,
  title = 	 {{FedPAQ:} {A} Communication-Efficient Federated Learning Method with Periodic Averaging and Quantization},
  author =     {Reisizadeh, Amirhossein and Mokhtari, Aryan and Hassani, Hamed and Jadbabaie, Ali and Pedarsani, Ramtin},
  journal = 	 {Proceedings of the International Conference on Artificial Intelligence and Statistics},
  pages = 	 {2021--2031},
  year = 	 {2020},
  volume = 	 {108}}

@article{andrew_2021_dp,
  title={Differentially private learning with adaptive clipping},
  author={Andrew, Galen and Thakkar, Om and McMahan, Brendan and Ramaswamy, Swaroop},
  journal={Proceedings of the International Conference on Advances in Neural Information Processing Systems},
  volume={34},
  pages={17455--17466},
  year={2021}
}

@article{chaudhuri_2011_dp,
  title={Differentially private empirical risk minimization},
  author={Chaudhuri, Kamalika and Monteleoni, Claire and Sarwate, Anand D},
  journal={Journal of Machine Learning Research},
  volume={12},
  number={3},
  year={2011}
}
\newpage
\appendix
\section{Lyapunov Analysis under Stochastic Disturbances}

\begin{lemma} \label{lemma:alternative_main} Let Assumptions~\ref{ass:uni_Lipschitz_f},  \ref{ass:convergence}, \ref{ass:smooth_flow}, \ref{ass:vanish_hessian_g} and \ref{ass:smooth_d} hold.
Then, there exists a Lyapunov function $V:\mathbb{N}\times \mathbb{R}^d \rightarrow [0,\infty)$ defined by
\begin{align}
    {V}(k,\xi)=\sum_{k'=0}^{M}\Phi(k,k')\dm\left(\phi(k',k,\xi),x_\ast\right)\label{eqn:Vtilde_def}
\end{align} where $\Phi$ is defined in \eqref{eqn:phi_definition} and $M$ is a sufficiently large integer such that Assumption~\ref{ass:convergence} holds.
This function satisfies the following inequalities,
\begin{align}\label{eqn:Vtilde_bound}
   \dm\left(\xi,x_\ast\right)\leq {V}(k,\xi)&\leq  \bar{c}_0  \dm\left(\xi,x_\ast\right)\\
    {V}(k+1,\phi(1,k,\xi)) &\leq \tau(k) {V}(k,\xi),\label{eqn:Vtilde_decrease}
\end{align} for all $k \geq 0$ and all $\xi \in \mathcal{S} $, where $S\subset \mathbb{R}^d$ is a compact set, and $\bar{c}_0$ is a constant.
In addition, for the states of the disturbed system $g_k(z_k,n_k)$ with zero-mean stochastic noise $n_k$, $\mathbb{E}\{|n_k|^2\}\leq\sigma_n^2$, there exists a constant $L_H$ such that
\begin{align}
    \mathbb{E}\left\{{V}(k\!+\!1,g_k(z_{k},n_k)) \right\} \leq \tau(k) {V}(k,z_k)+\frac{1}{2}L_H L_{\mathrm{e}k}^2{\sigma_n^2},\label{eqn:V_tilde_pertdecrease_inlemma}
\end{align}for all $z_k \in \mathcal{S}$, where $L_{\mathrm{e}k}>0$ is defined in Assumption~\ref{ass:vanish_hessian_g}. %
\end{lemma}

\begin{proof}
The comparison inequalities in \eqref{eqn:Vtilde_bound} follow directly from \eqref{eqn:stability} and the definition of $V$ in \eqref{eqn:Vtilde_def}. Indeed, we establish the upper bound as\begin{align*}
        {V}(k,\xi) &=\sum_{k'=0}^{M}\Phi(k,k')\dm\left(\phi(k',k,\xi),x_\ast\right)\leq M c_0\dm\left(\xi,x_\ast\right)=\bar{c}_0\dm\left(\xi,x_\ast\right).
\end{align*} 
The lower bound follows from the first term in the sum,
\begin{align*}
    {V}(k,\xi) &=\sum_{k'=0}^{M}\Phi(k,k')\dm\left(\phi(k',k,\xi),x_\ast\right)\geq \dm\left(\xi,x_\ast\right).
\end{align*}
Next, we analyze the decrease condition,
\begin{align*}
     {V}(k+1,\phi(1,k,\xi)) &=\sum_{k'=0}^{M}\Phi(k+1,k')\dm\left(\phi(k',k+1,\phi(1,k,\xi)),x_\ast\right)\\
     &=\sum_{k'=0}^{M}\Phi(k+1,k')\dm\left(\phi(k'+1,k,\xi),x_\ast\right)\\
      &=\sum_{k'=0}^{M}\Phi(k,k'+1)\tau(k)\dm\left(\phi(k'+1,k,\xi),x_\ast\right)\\
     &=\tau(k)V(k,\xi)+\tau(k)\left(\Phi(k,M\!+\!1)\dm\left(\phi(M\!+\!1,k,\xi),x_\ast\right)-\dm\left(\xi,x_\ast\right)\right)
     \\
     &\leq \tau(k) {V}(k,\xi),
\end{align*}
where the last inequality follows from \eqref{eqn:ass_2}. 
Specifically, by Assumption~\ref{ass:convergence}, there exists an integer $K>0$ such that for all $M\geq K$,
\begin{align*}
   \dm\left(\phi(M+1,k,\xi),x_\ast\right)
   \leq(\Phi(k,M+1))^{-1}\dm\left(\xi,x_\ast\right),
\end{align*} which implies the last term in brackets is nonpositive for $M\geq K$.
Hence, choosing any finite $M\geq K$ yields \eqref{eqn:Vtilde_decrease}.

\newpage
For the states of the disturbed system, since $ \tfrac{\partial^2 g_k}{\partial n^2} = 0 $ by Assumption~\ref{ass:vanish_hessian_g}, we get, $$ g_k(z_k, n) = g_k(z_k, 0) + \left.\frac{\partial g_k}{\partial n}\right|_{n=0} n. $$ Thus, $ V(k+1, g_k(z_k, n)) $ is twice differentiable, and its Taylor expansion around $ n = 0 $ is
\begin{align}\begin{aligned}
    V(k\!+\!1,g_k(z_k,n))
    \!&= V(k\!+\!1,g_k(z_k,0))
   \\&+ \nabla_{x} V(k+1,g_k(z_k,0))^\top
      \left.\frac{\partial g_k(z_k,\bar n)}{\partial \bar n}\right|_{\bar n=0} \!\!n \\
    &+\! \frac{1}{2}  n^{\!\!\top}\!\!\left(\!\!
      \left.\!\frac{\partial g_k(z_k,\bar n)}{\partial \bar n}\!\right|_{\bar n=\bar\varepsilon}^{\!\top}\!\!\!
      \left.\frac{\partial^2 V\!(k\!+\!1,\bar{x})}{\partial x^2}\!\right|_{\bar x=g_k(z_k,\bar\varepsilon)}\!\!\!\!\!
      \left.\frac{\partial g_k(z_k,\bar{n})}{\partial \bar n}\!\right|_{\bar n=\bar\varepsilon}\!
      \right)\!n.
\end{aligned}\label{eqn:V_tilde_taylor}
\end{align}
Since the Hessian of $g_k$ with respect to $n$ vanishes by Assumption~\ref{ass:vanish_hessian_g}, the remainder depends only on the Hessian of ${V}$. The Hessian of $V$ can be computed as 
\begin{align*}
&\begin{aligned}
 \frac{\partial^2 {V}(k,\xi)}{\partial \xi^2}   
        &=\sum_{k'=0}^{M}\Phi(k,k')\frac{\partial^2 \dm\left(\phi(k',k,\xi),x_\ast\right)}{\partial \xi^2}  \\
&= \sum_{k'=0}^{M} \Phi(k,k') \Bigg[ 
\left(\frac{\partial \phi(k',k,\xi)}{\partial \xi}\right)^\top 
\left.\frac{\partial^2 \dm\left(x,x_\ast\right)}{\partial x^2}\right|_{x=\phi(k',k,\xi)} 
\left(\frac{\partial \phi(k',k,\xi)}{\partial \xi}\right)
\Bigg. \\
&\qquad\qquad\qquad\qquad \Bigg. 
+  C(k',k,\xi)
\Bigg],\end{aligned}\\
&\text{where}\quad[C(k',k,\xi)]_{ij}=\sum_{p=1}^{d} \left( \frac{\partial^2 \phi_p(k',k,\xi)}{\partial \xi_i\partial\xi_j} \right)^\top 
\left.\frac{\partial \dm\left(x,x_\ast\right)}{\partial x_p}\right|_{x=\phi(k',k,\xi)}.
\end{align*}
By Assumption~\ref{ass:smooth_flow} and the definition of $\Phi$, we have $\Phi(k,k')\le\tau(0)^{-k'}$, and combined with Assumption~\ref{ass:uni_Lipschitz_f}, this implies
\begin{align}
    \left|\frac{\partial^2 {V}(k,\xi)}{\partial \xi^2}  \right|\leq M\tau(0)^{-M}(L_{H_\dm~} L_f^{2M}+ L_{\dm~} (L_HL_f^M(L_f+M)) ),
\end{align}where $L_{H_\dm~}$ is defined in Assumption~\ref{ass:smooth_d}. 
We denote the right-hand side by a constant $L_H>0$, that is,
$$L_H:= M\tau(0)^{-M}(L_{H_\dm~} L_f^{2M}+ L_{\dm~} L_f^M(L_f+M) ).$$
Substituting this bound and \eqref{eqn:Vtilde_decrease} into \eqref{eqn:V_tilde_taylor} and taking the expectation with respect to $n$ with $\mathbb{E}\{n\}=0$ yields
\begin{align}
        \mathbb{E}\left\{{V}(k+1,g_k(z_{k},n_k)) \right\} \leq \tau(k) {V}(k,z_k)+\frac{1}{2}L_H L_{\mathrm{e}k}^2{\sigma_n^2}.
\end{align}

\end{proof}

\end{document}